\documentclass[letterpaper, 10 pt, conference]{ieeeconf}
\IEEEoverridecommandlockouts 
\usepackage{hyperref, url, amsmath, amssymb, booktabs, graphicx, bm, color, algorithm, tabularx}
\usepackage[noend]{algpseudocode}
\graphicspath{{images/}}

\DeclareMathOperator*{\argmin}{\arg\min}
\newcolumntype{Y}{>{\centering\arraybackslash}X}

\title{\LARGE \bf Deep Haptic Model Predictive Control for Robot-Assisted Dressing}

\author{Zackory Erickson, Henry M. Clever, Greg Turk, C. Karen Liu, and Charles C. Kemp
\thanks{Zackory Erickson, Henry M. Clever, and Charles C. Kemp are with the Healthcare Robotics Lab, Georgia Institute of Technology, Atlanta, GA., USA.}%
\thanks{C. Karen Liu and Greg Turk are with the School of Interactive Computing, Georgia Institute of Technology, Atlanta, GA., USA.}%
\thanks{Zackory Erickson is the corresponding author {\tt\footnotesize zackory@gatech.edu}.}%
}

\begin{document}

\maketitle
\thispagestyle{empty}
\pagestyle{empty}

\begin{abstract}
Robot-assisted dressing offers an opportunity to benefit the lives of many people with disabilities, such as some older adults. However, robots currently lack common sense about the physical implications of their actions on people. The physical implications of dressing are complicated by non-rigid garments, which can result in a robot indirectly applying high forces to a person's body. We present a deep recurrent model that, when given a proposed action by the robot, predicts the forces a garment will apply to a person's body. We also show that a robot can provide better dressing assistance by using this model with model predictive control. The predictions made by our model only use haptic and kinematic observations from the robot's end effector, which are readily attainable. Collecting training data from real world physical human-robot interaction can be time consuming, costly, and put people at risk. Instead, we train our predictive model using data collected in an entirely self-supervised fashion from a physics-based simulation. We evaluated our approach with a PR2 robot that attempted to pull a hospital gown onto the arms of 10 human participants. With a 0.2s prediction horizon, our controller succeeded at high rates and lowered applied force while navigating the garment around a persons fist and elbow without getting caught. Shorter prediction horizons resulted in significantly reduced performance with the sleeve catching on the participants' fists and elbows, demonstrating the value of our model's predictions. These behaviors of mitigating catches emerged from our deep predictive model and the controller objective function, which primarily penalizes high forces.
\end{abstract}

\section{Introduction}
\label{sec:intro}

Robotic assistance presents an opportunity to benefit the lives of many people with disabilities, such as some older adults. However, robots currently lack common sense about the physical implications of their actions on people when providing assistance.
Assistance with dressing can improve a person's quality of life by increasing his or her independence and privacy. Yet, dressing presents further difficulties for robots due to the complexities that arise when manipulating fabric garments around people.

Model predictive control (MPC) enables robots to account for errors and replan actions in real time when interacting in dynamic environments. For example, MPC has found success in several contexts such as obstacle avoidance and object manipulation~\cite{finn2017deep,jain2013reaching,lenz2015learning}.
However, these existing robotic controllers do not take into consideration the physical implications of a robot's actions on a person during physical human-robot interaction. This is especially true during robot-assisted dressing in which a robot may never make direct physical contact with a person, but instead apply force onto the person through an intermediary non-rigid garment. Yet, robots could greatly benefit from predicting the physical implications of their actions when interacting with people.

\begin{figure}
\centering
\includegraphics[width=0.48\textwidth, trim={15cm 8cm 6cm 1cm}, clip]{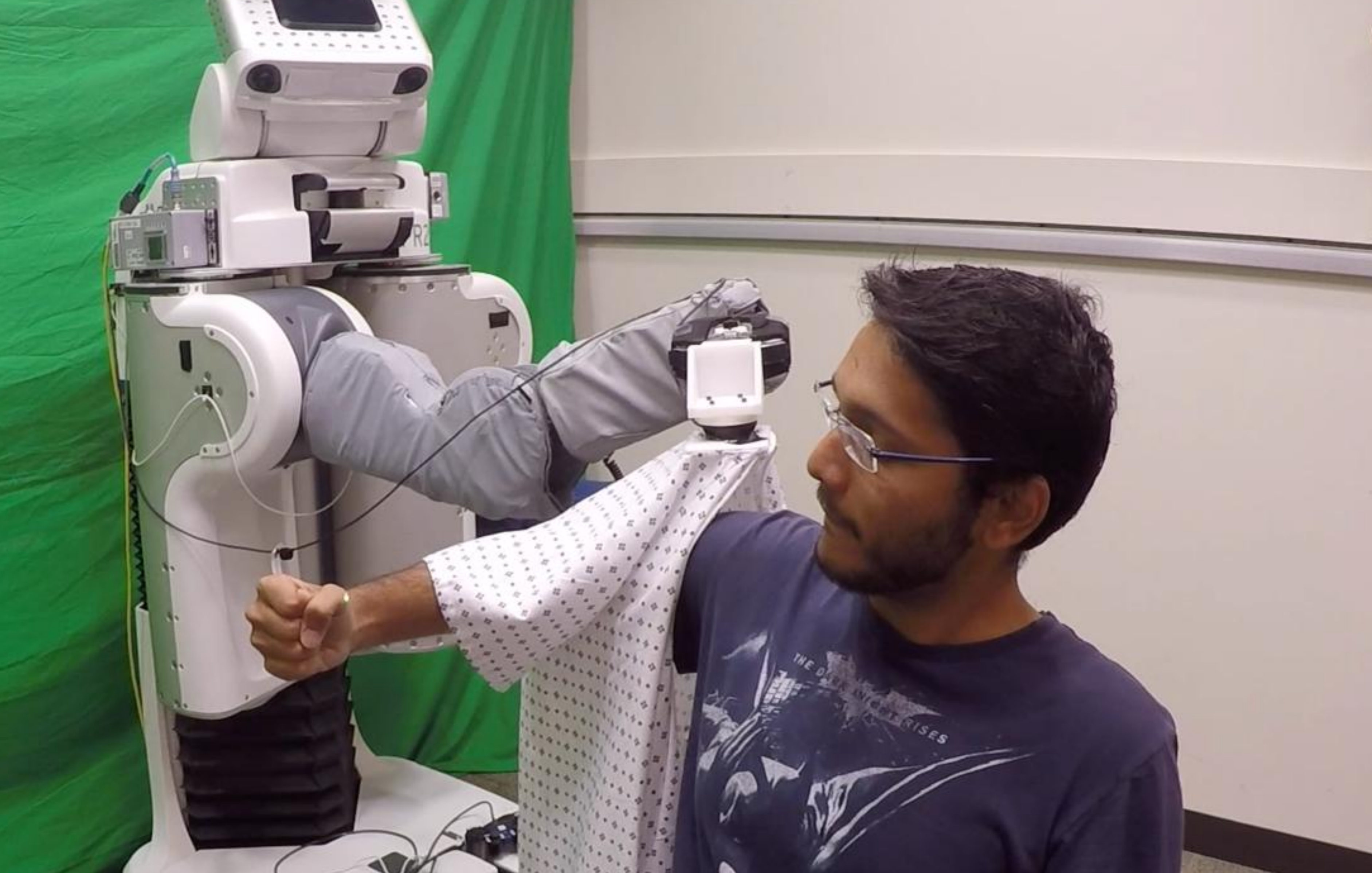}
\vspace{-0.4cm}
\caption{\label{fig:intro}Using our approach, a PR2 pulls a hospital gown onto a participant's arm by minimizing the predicted forces applied to the person's body.}
\vspace{-0.4cm}
\end{figure}

In this paper, we propose a Deep Haptic MPC approach that allows a robot to minimize the predicted force it applies to a person during robotic assistance that requires physical contact.
We train a recurrent model that consists of both an estimator and predictor network in order to predict the forces applied onto a person, and
we detail the benefits of this approach in Section~\ref{sec:simmodel}. The estimator outputs the location and magnitude of forces applied to a person's body given haptic sensory observations from a robot's end effector. The predictor outputs future haptic observations given a proposed action. Together, these two networks allow a robot to determine the physical implications of its actions by predicting how future actions will exert forces onto a person's body.
We demonstrate our approach on a real robotic system that assisted 10 human participants in pulling a hospital gown onto a person's right arm, as seen in Fig.~\ref{fig:intro}.


We train our model on data generated entirely in a physics-based simulation, allowing us to quickly collect thousands of diverse training sequences that would otherwise be dangerous or infeasible to collect on real robotic systems that physically interact with people. Our simulated robot can make mistakes, explore new approaches for interaction, and investigate error conditions without putting real people at risk.

These training data are generated in a self-supervised fashion, without a reward function or specified goal. Once training is complete, we define an objective function that enables our controller to prioritize future actions that minimize the predicted force applied to a person during dressing. Since our model is trained without a predefined reward function, we can redefine the objective function without retraining the model. 
We further compare dressing results for various time horizons with MPC and observe emergent behaviors as the prediction horizon increases.

The key contribution of this paper is to demonstrate that a deep recurrent model over haptic and kinematic measurements can be used by real robotic systems to predict the physical implications of future actions and lower the forces applied to a person during robot-assisted dressing. We show that this model can be trained in simulation and applied to a real robotic task of pulling a garment onto a person's arm. By combining our learning-based model with MPC, we observe emergent behaviors that result in the robot navigating a garment up a person's entire arm.

\section{Related Work}
\label{sec:related_work}


\subsection{Robot-Assisted Dressing and Force Estimation}

Several robotic dressing approaches have relied on visual systems to estimate a person's pose and the state of a garment. For example, Koganti et al.~\cite{koganti2017bayesian} used RGB-D and motion capture data to estimate the topological relationship between a person's body and a garment. Klee et al.~\cite{klee2015personalized} visually detected a person's pose which was used by a Baxter robot to assist in putting on a hat.
Pignat et al.~\cite{pignat2017learning} tracked a person's hand movement in real time using an AR tag. The researchers then used a Baxter robot to pull one sleeve of a jacket onto a person's arm. Unlike this body of work, our approach does not rely on visual observations, but is instead able to fully dress a person's arm using only haptic and kinematic measurements obtained at the robot's end effector.

Several researchers have similarly explored haptic sensing within the context of robot-assisted dressing. Gao et al.~\cite{gao2016iterative} proposed a force feedback control approach that allowed a Baxter robot to assist in dressing a sleeveless jacket. Kapusta et al.~\cite{kapustadata} explored how haptic observations at a robot's end effector can be coupled with an HMM to predict the future outcome of a dressing task. Yamazaki et al.~\cite{yamazaki2014bottom} described a failure detection approach for robot-assisted dressing that leveraged force data while assisting participants in pulling up pants. Instead, our work demonstrates that haptic sensing and learning can be used to predict the physical implications of a robot's future actions when assisting people. When coupled with MPC, we show that these predictions also enable a robot to replan its actions in real time during robotic assistance.

In prior work~\cite{erickson2017does} we presented an LSTM model trained in simulation to estimate the forces applied onto a simulated arm and leg during robot-assisted dressing tasks. The estimator we present in this paper uses a similar network architecture except we also provide end effector position and yaw rotation measurements to the model so that our PR2 can navigate around a person's elbow. We pair this estimator with a predictor and evaluate a PR2's ability to predict the physical implications of its actions during assistance.

\subsection{Model Predictive Control}

Model predictive control has found success in several robotics domains. Some examples include aerial control vehicles~\cite{abbeel2010autonomous,bellingham2002receding} and robot locomotion~\cite{erez2012infinite,wieber2006trajectory}.  This work has similarity to~\cite{chow2016robotic,jain2013reaching,dominici2014model}, using haptic information as a model input for control in the manipulation domain.  Prior robotics research has used analytical models for MPC ~\cite{chow2016robotic,duchaine2007computationally,dominici2014model,jain2013reaching}, whereas we employ a learning-based model as in~\cite{fu2016one,nguyen2011model,lenz2015learning}. 

Many past works have relied on vision-based approaches for robotic control with MPC. 
Finn and Levine~\cite{finn2017deep} combined a predictive model of image observations with MPC for nonprehensile pushing tasks. Watter et al.~\cite{watter2015embed} presented a learning-based control method for non-linear dynamical systems using raw pixel images. Boots et al.~\cite{boots2014learning} learned a predictive model that generates RGB-D images of a robot arm moving in free space.
In comparison to these vision-based methods, our learned model uses only haptic and kinematic information.
Chow et al.~\cite{chow2016robotic} leveraged haptic observations with MPC to assist in reposition a person's limbs in simulation.
Lenz et al.~\cite{lenz2015learning} learned material properties for cutting various foods with a PR2, but rely on joint torques for haptic feedback, which have a lower dimensionality and accuracy than the 6-DoF discrete force/torque sensor in our system. In addition, Jain et al.~\cite{jain2013reaching} showed how a robot arm can reach into cluttered spaces using haptic sensing skin.

Learning models with neural networks for robot control is common throughout many robotic control approaches~\cite{nguyen2011model}.
Lenz et al.~\cite{lenz2015learning} used a recurrent model with MPC and demonstrated their approach on a PR2 that learned deep latent material properties by performing 1,488 cuts across 20 foods.
Finn and Levine~\cite{finn2017deep} combined a deep predictive model of image sequences with MPC and trained their model on 50,000 pushing attempts of objects using 7-DoF manipulators.
Unlike these approaches, our model is trained entirely in simulation, which presents several benefits for physical human-robot interaction, as discussed in Section~\ref{sec:simmodel}.
  
Fu et al.~\cite{fu2016one} used model-based reinforcement learning in simulation, where a PR2 learned to manipulate rigid objects with MPC.
Unlike reinforcement learning, our method does not require a reward function during training, which allows us to decouple the objective function from the learned model.
Furthermore, we show that our learning-based model can enable a real PR2 to predict the physical implications of its actions when assisting human participants with dressing.

\section{Simulation and Model Training}
\label{sec:simmodel}

\begin{figure}
\centering
\includegraphics[width=0.48\textwidth, trim={3cm 0cm 0cm 0cm}, clip]{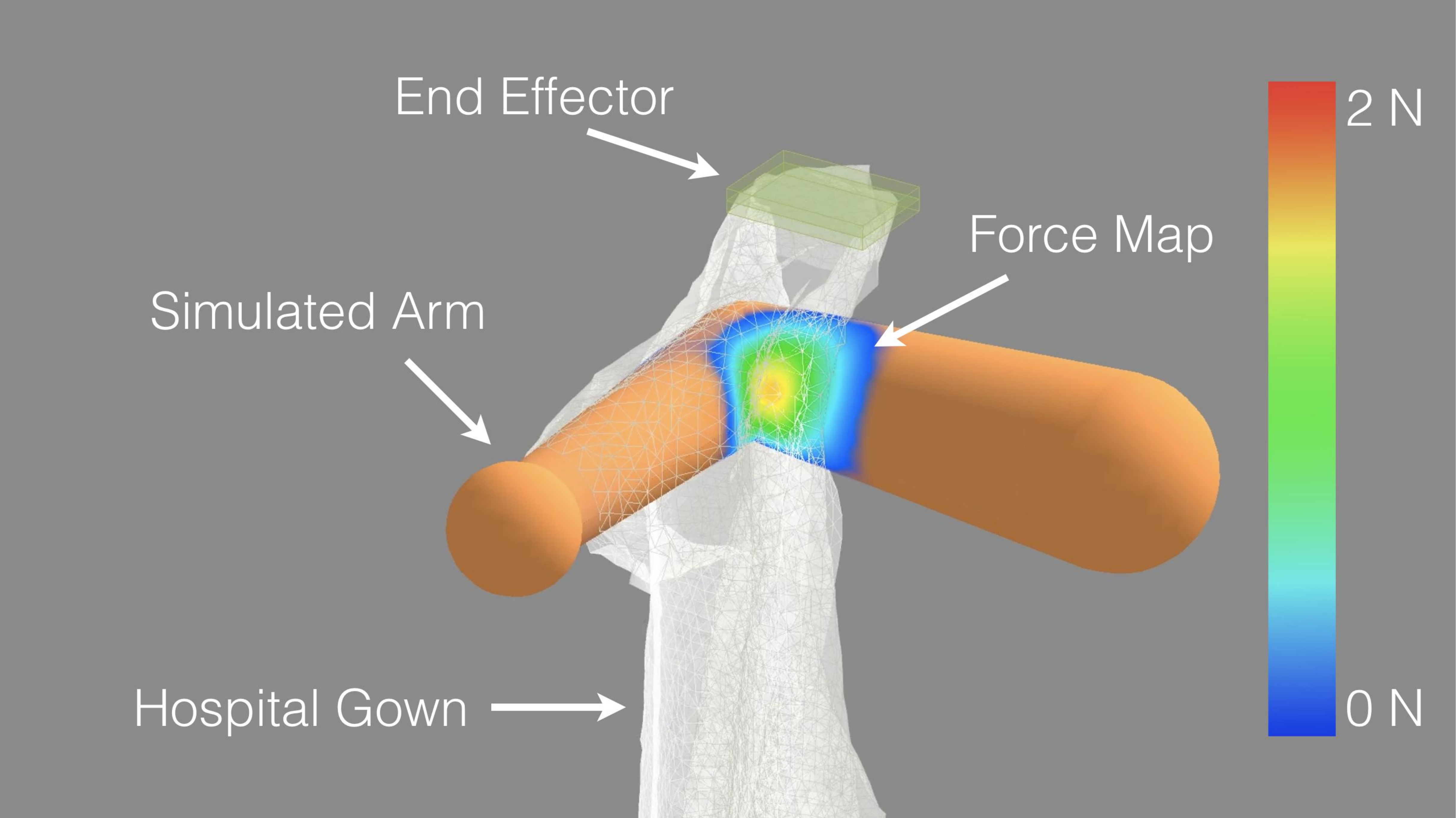}
\vspace{-0.4cm}
\caption{\label{fig:simulation}The physics-based simulation environment used to generate a training set for our model. The simulation records force, torque, position, and velocity from the robot's end effector, as well as all forces applied onto the arm by the fabric mesh gown.}
\vspace{-0.4cm}
\end{figure}

To perform deep haptic MPC that considers forces applied to a person, our model consists of two recurrent neural networks trained on a dataset of simulated robot-assisted dressing trials. Here we introduce notation and we provide a brief description of the model, simulation, and data collection process. Our dataset consists of 10,800 dressing trials generated in a simulated robot-assisted dressing environment 
presented in prior work~\cite{erickson2017does}. As shown in Fig.~\ref{fig:simulation}, this physics-based simulation consists of a robotic end effector that pulls a hospital gown onto a simulated human arm. The colored fields along the arm represent a force map that encompasses a set of force magnitudes applied at specific locations on the body. Several advantages arise from collecting data with a physics-based simulation. First, we can easily parallelize data collection to collect thousands of dressing experiences in a few hours. We can also test anomalous scenarios that may be infeasible or dangerous to test with real people, such as cloth getting caught on a body part. These anomalous conditions could be especially valuable for a robot, so that it can learn to mitigate potentially harmful consequences. 
Finally, we can calculate the location and magnitude of all forces applied to a person by a clothing garment within simulation, something that is highly challenging in the real world.

During data collection, the simulated robotic end effector attempts to pull the sleeve of a hospital gown onto the person's arm. The simulator randomly selects a starting position near the arm and movement velocity for the end effector prior to each trial. During a dressing trial, the simulation iteratively selects a new random action for the robot's end effector at each time step. In doing so, our model learns about diverse situations for a garment to make contact with a person's arm. 
We represent actions in the fixed coordinate frame of the robot's torso, and actions consist of a 3D velocity for the end effector and a change in yaw rotation around an axis parallel to gravity, i.e. $\bm{a}=(v_x, v_y, v_z, \Delta\psi)$. The simulation selects new actions at 5 Hz and records sensor measurements at 100 Hz. Measurements $\bm{x}_t = (\bm{\rho}, \bm{v}, \psi, \bm{f}^r, \bm{\tau}^r) \in \mathbb{R}^{13}$, at time $t$, include the 3D position $\bm{\rho}$, 3D velocity $\bm{v}$, and yaw rotation $\psi$ of the end effector, and the 3D forces $\bm{f}^r$ and torques $\bm{\tau}^r$ applied at the robot's end effector by the garment. We record all forces applied to the human's arm, which occur when a vertex on the fabric mesh makes contact with the simulated arm. We construct a force map, as shown in Fig.~\ref{fig:simulation}, by mapping these applied forces to a discrete set of fixed points (taxels) spaced across the surface of the arm. \cite{erickson2017does} provides further details of this mapping procedure and force map definition. In this work, we use 37 taxels distributed across the fist, forearm, and upper arm.

As Yu et al.~\cite{yu2017haptic} proposed, we used Covariance Matrix Adaptation Evolution Strategy (CMA-ES)~\cite{hansen2016cma} to optimize the parameters of our simulator with respect to data collected from a real robotic system that assisted human participants in pulling on a hospital gown. Some of these parameters include garment stretch, stiffness, shear forces, and friction. Because of this optimization, the force and torque measurements in simulation align closely to those observed in the real world. However, the simulated end effector performs exact movements, whereas the motion trajectory of a PR2's end effector often includes noise due to the compliant nature of the arms. To account for this, we added a small amount of uniformly sampled noise, $\xi\in[-0.8, 0.8]$ mm/s, to each component of the end effector's velocity at every time step in the simulation. During model training, this also serves as a form of regularization to help mitigate overfitting to the position and velocity measurements from simulation. 



We leverage a pair of recurrent networks to predict the forces applied to a person given a sequence of proposed robot actions. We define a predictor $G(\bm{x}_{1:t}, \bm{a}_{t+1:t+H_p})$, which predicts a sequence of future end effector haptic measurements, $\bm{\hat{x}}_{t+1:t+H_p}$, that result from the robot executing actions $\bm{a}_{t+1:t+H_p}$ over a prediction horizon $H_p$. We then use an estimator, $F(\bm{x}_{1:t}, \bm{\hat{x}}_{t+1:t+H_p})$, that estimates the forces, $\bm{f}_{t+H_p}$, applied to a person at time $t+H_p$ given all prior measurements $\bm{x}_{1:t}$, and the predicted measurements, $\bm{\hat{x}}_{t+1:t+H_p}$.
We can predict future force maps by composing the estimator and predictor, $F \circ G = F(\bm{x}_{1:t}, G(\bm{x}_{1:t}, \bm{a}_{t+1:t+H_p}))$, wherein we estimate force maps given predicted haptic measurements. Furthermore, we can make predictions beyond time $t+H_p$ by feeding the predicted measurements $\bm{\hat{x}}_{t+1:t+H_p}$ back into $G$ along with an action sequence $\bm{a}_{t+H_p+1:t+2H_p}$. Thus, $\bm{\hat{x}}_{t+H_p+1:t+2H_p}$ can be predicted via $G(\{\bm{x}_{1:t},\bm{\hat{x}}_{t+1:t+H_p}\},\bm{a}_{t+H_p+1:t+2H_p})$.

Although these two networks could be merged, there are several advantages to a split architecture. First, this setup allows for additional flexibility in that a new predictor can be learned without impacting the accuracy of force map estimation, or vise versa. Furthermore, we are able to run these two networks at different frequencies which is beneficial during real time use. We run the estimator at 100 Hz as this results in greater accuracy and resolution for force map estimates. However, since we update the robot's action at 5~Hz, we need only make predictions at a 5~Hz rate. In total, the estimation model receives 20 measurements for each step of the prediction model. This difference in frequencies was crucial for real-time implementation since prediction is a computationally demanding task for each candidate action.

This approach also presents several advantages over formulating a reinforcement learning problem and solving for a policy. Our objective function is decoupled from the learned model, thus the objective can be redefined for different dressing tasks without retraining the model. Also, the data we collect for training the estimator can be reused to train the predictor, whereas model-free reinforcement learning methods require further data collection and new sets of rollouts from the evolving policy after training an estimator.

\begin{figure}
\centering
\includegraphics[width=0.48\textwidth, trim={0cm 0cm 0cm 0cm}, clip]{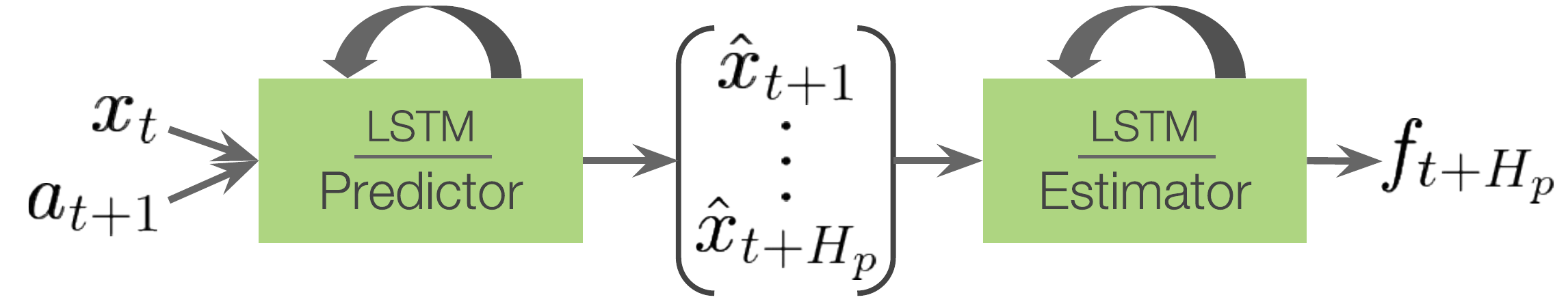}
\vspace{-0.4cm}
\caption{\label{fig:lstm}Our estimator and predictor LSTM networks with all associated inputs and outputs.}
\vspace{-0.1cm}
\end{figure}

\begin{figure}
\centering
\includegraphics[width=0.48\textwidth, trim={0cm 0cm 0cm 0cm}, clip]{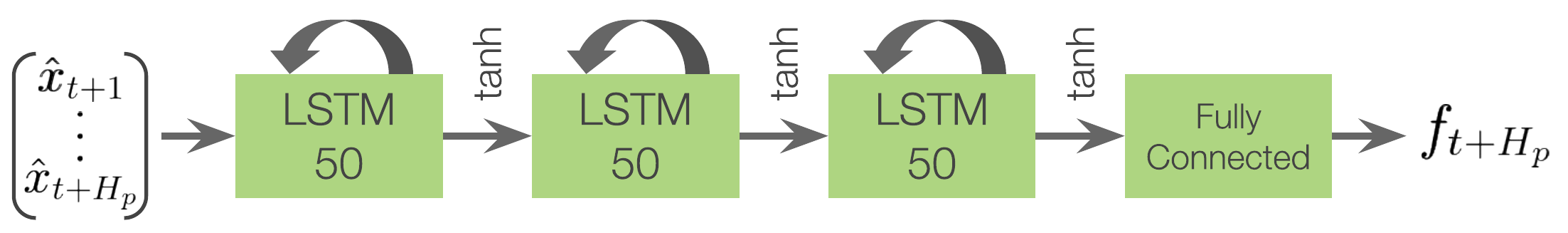}
\vspace{-0.4cm}
\caption{\label{fig:lstm_network}Network architecture for the estimator.}
\vspace{-0.1cm}
\end{figure}


As shown in Fig.~\ref{fig:lstm}, our model uses LSTMs to estimate force maps and predict future measurements. Each recurrent model consists of three LSTM layers with 50 recurrent cells and a \textit{tanh} activation. The final output layer is fully connected with a linear activation. Fig.~\ref{fig:lstm_network} shows this network architecture for our estimator model. The predictor uses the same architecture, but with different input/output. Note that if the robot maintains a constant action throughout the entire prediction horizon, as is the case in our work, a sequence of identical actions, $\bm{a}_{t+1:t+H_p}$, can be collapsed down to a single action, $\bm{a}_{t+1}$. Because of this, our predictor outputs a sequence of measurements $\bm{\hat{x}}_{t+1:t+H_p}$ given a single action, $\bm{a}_{t+1}$, and measurement, $\bm{x}_t$. We use $H_p=20$, which aligns with the 5~Hz rate used for predictions when the difference between time steps is 0.01s. Ideally, the predictor would evaluate sequences of actions that vary over time. However, our experiments showed that using the same action over the entire prediction horizon was computationally tractable and worked well in practice for physical human-robot interaction. Related literature has also found a 5~Hz action replanning rate to be computationally feasible for MPC on real robots~\cite{finn2017deep}.

\section{Model Predictive Control}
\label{sec:mpc}

\begin{figure}
\centering
\includegraphics[width=0.48\textwidth, trim={15cm 12cm 10cm 5cm}, clip]{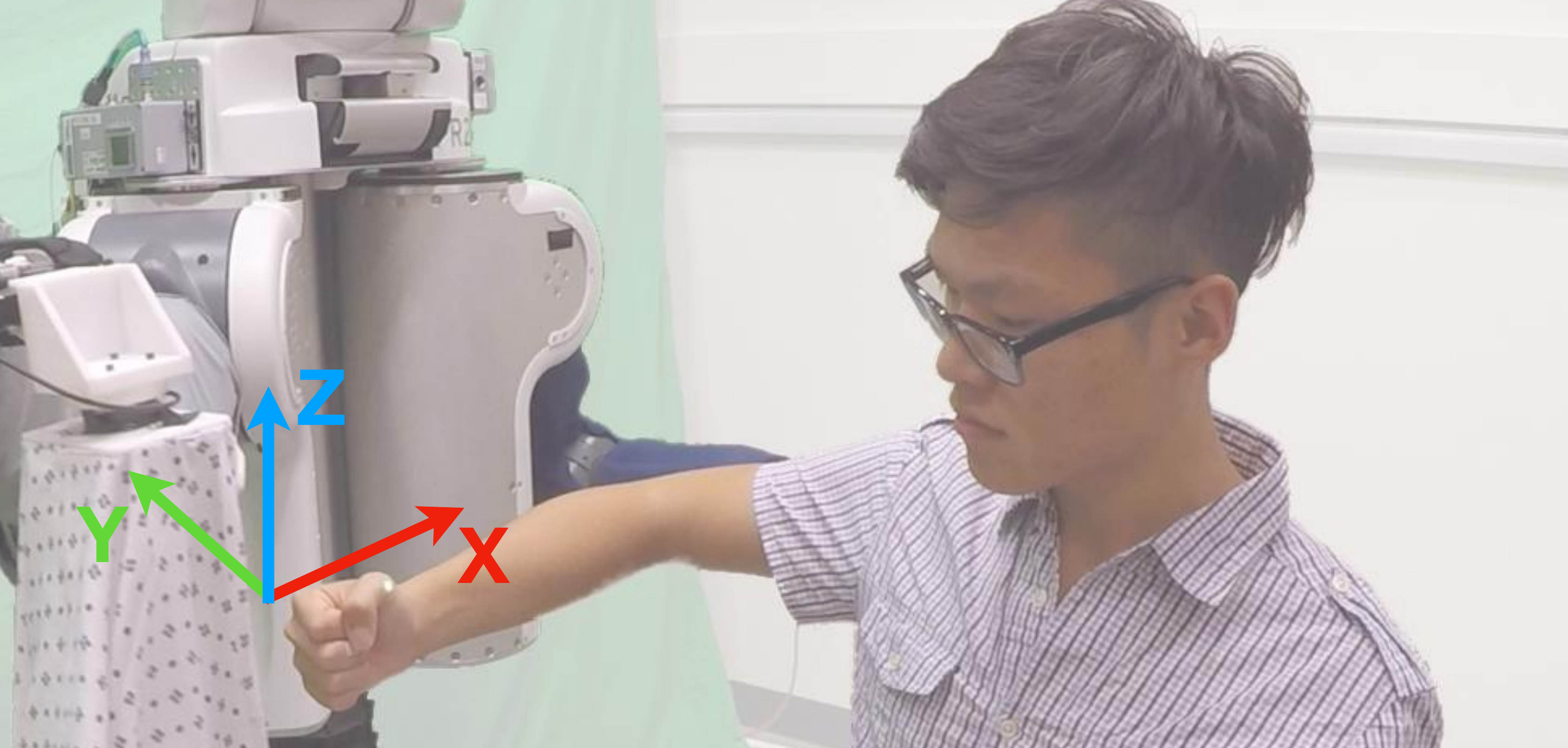}
\vspace{-0.4cm}
\caption{\label{fig:armpose}Initial configuration for the robot and a participant prior to the start of a dressing trial. We overlay the global coordinate axes used by the robot when replanning actions.}
\vspace{-0.4cm}
\end{figure}

Our system uses model predictive control (MPC) with our recurrent estimator and predictor to choose actions that minimize the predicted force applied to a person during physical assistance. Here we present the cost function that we used to encourage certain robot actions and we describe ways in which this function could be adapted to allow for personalized robotic assistance. In addition, we present our MPC method for replanning actions, which involves predicting applied forces for a set of candidate robot actions.

We define a cost function leading to lower forces applied to a person during dressing assistance. The cost function input includes the current and prior measurements, $\bm{x}_{1:t}$ and a sequence of candidate actions, $\bm{a}_{t+1:t+H_p}$. In addition to penalizing large forces applied on the person's body, the cost function encourages forward moving end effector actions and penalizes yaw rotations, represented by three weighted terms:

\vspace{-0.4cm}
\begin{equation}
\label{eq:cost}
\begin{split}
J(\bm{x}_{1:t},\bm{a}_{t+1:t+H_p}) &= w_1\left\Vert F(\bm{x}_{1:t}, G(\bm{x}_{1:t}, \bm{a}_{t+1:t+H_p}))\right\Vert^2_1 \\
&\quad - w_2 \sum\limits_{j=t+1}^{t+H_p} \bm{\bar{d}} \cdot \bm{a}_{j,v} \\
&\quad + w_3 \sum\limits_{j=t+1}^{t+H_p} |a_{j,\psi}|
\end{split} \raisetag{3\baselineskip}
\end{equation}
where $\bm{a}_j=(v_x, v_y, v_z, \Delta\psi)$ represents a candidate action, $\bm{a}_{j,v}$ represents the 3-axis velocity components of the action, $a_{j,\Delta\psi}$ represents the yaw rotation component of the action, $\bm{\bar{d}}=(1, 0, 0)$ depicts a forward moving action, and $w_1, w_2, w_3$ are constant weights set based on the importance of making task progress versus keeping forces low. The first term, $\left\Vert F(\bm{x}_{1:t}, G(\bm{x}_{1:t}, \bm{a}_{t+1:t+H_p}))\right\Vert_1$, represents the $L_1$ norm of all predicted forces, $\bm{f}_{t+H_p}$, at the 37 taxels along a person's arm at time $t+H_p$. We square this term to reduce the influence of small forces that occur at the beginning of dressing. This is supported by the notion that small forces are unlikely to cause issues during assistance~\cite{jain2013reaching}. However, as more force is applied to a person's arm, this term becomes the dominating factor for selecting which action the robot will execute. For various applications, this $L_1$ norm term may also be modified to focus on certain body joints, e.g. only minimizing force around the hand and wrist, rather than the entire arm. The second term, $\bm{\bar{d}} \cdot \bm{a}_{j,v}$, rewards actions that move in a forward direction along the +X global coordinate axis, or approximately the central axis of a person's forearm, as shown in Fig.~\ref{fig:armpose}. The last term, $|a_{j,\psi}|$, penalizes actions that perform a yaw rotation. Without these last two terms, the optimal action to minimize cost is sometimes an action that performs no movement. Depending on the task, the terms in Equation (\ref{eq:cost}) may also be combined via a nonlinear function to support a variety of complex behaviors. From our experiments described in Section~\ref{sec:evaluation}, we observe that this simple cost function can lead to emergent behaviors in which the robot can navigate a garment up a person's entire arm. 

We update the robot's action by selecting the sequence of actions that minimize Equation (\ref{eq:cost}).
This can be denoted as,
\begin{equation}
\label{eq:update}
\bm{a}^*_{t+1:t+H_p} = \underset{\bm{a}_{t+1:t+H_p}}\argmin \ J(\bm{x}_{1:t}, \bm{a}_{t+1:t+H_p}).
\end{equation}

\begin{algorithm}[t]
\caption{Deep Haptic Model Predictive Control}\label{alg:mpc}
\begin{algorithmic}[1]
\State \textbf{input:} estimation model $F$, prediction model $G$, cost function $J$, time steps between predictions $\tau_p$.
\State $t \gets 0$.
\While{force $<$ 10 N and joint limits not reached}
\State Observe new sensor measurement $\bm{x}_t$.
\If{$t \bmod \tau_p = 0$}
\State Initialize $N$ action sequences $\{\bm{a}_{t+1:t+H_p}^{(n)}\}$.
\State Select $\bm{a}_{t+1:t+H_p}^*$ using Equation (\ref{eq:update}).
\State Execute $\bm{a}_{t+1:t+\tau_p}^*$.
\EndIf
\State $t \gets t+1$.
\EndWhile
\end{algorithmic}
\end{algorithm}

Algorithm~\ref{alg:mpc} presents our procedure for updating the robot's actions during robot-assisted dressing. At each time step $t$, we observe sensor measurements $x_t$. Every $\tau_p$ time steps, our controller chooses the actions $\bm{a}_{t+1:t+H_p}^*$ that minimize the cost function, based on a set of $N$ candidate action sequences, $\{\bm{a}_{t+1:t+H_p}^{(n)}\}$. In this work, we use $H_p=\tau_p=20$ and we initialize a fixed set of $N=28$ actions whose velocity lie within a hemisphere facing the +X global coordinate axis. Computing the cost for each action sequence involves predicting a sequence of future end effector measurements $\bm{\hat{x}}_{t+1:t+H_p}$ and feeding these measurements into the estimator, $F$, to estimate the force map at time $t+H_p$. We terminated a trial when the magnitude of forces measured at the robot's end effector exceeded 10 N, or the robot's arm reached its joint limits, which can occur when the arm fully extends to pull a garment onto a participant's shoulder.

This predictive control approach runs in real time on a PR2, using only the robot's on-board CPUs, and both our estimator and predictor can make predictions at $\sim$2~kHz. One limitation is that our model is constrained to relatively short horizon tasks. Notably, our system performed well even with short horizon planning. Computation time limits both the action replanning rate and the prediction horizon, yet our work leaves significant room for future improvements with GPUs, greater parallelization, and off-board computation. Additionally, in this work, we evaluate our model's predictive capabilities, so we restrict our controller from selecting actions that move "backwards". Future implementations could relax this for more freedom while replanning a trajectory.

\section{Evaluation}
\label{sec:evaluation}


We conducted experiments with 10 participants (2 female, 8 male) with approval from the Georgia Institute of Technology Institutional Review Board (IRB), and obtained informed consent from all participants.  We recruited able-bodied participants to meet the following inclusion/exclusion criteria: $\geq18$ years of age; have not been diagnosed with ALS or other forms of motor impairments; fluent in written and spoken English. Their ages ranged from 18 to 30 years. A video of our experiments can be found online\footnote{Video: \url{http://healthcare-robotics.com/haptic-mpc}}.

\begin{figure}
\centering
\includegraphics[width=0.48\textwidth, trim={10cm 8cm 7cm 1cm}, clip]{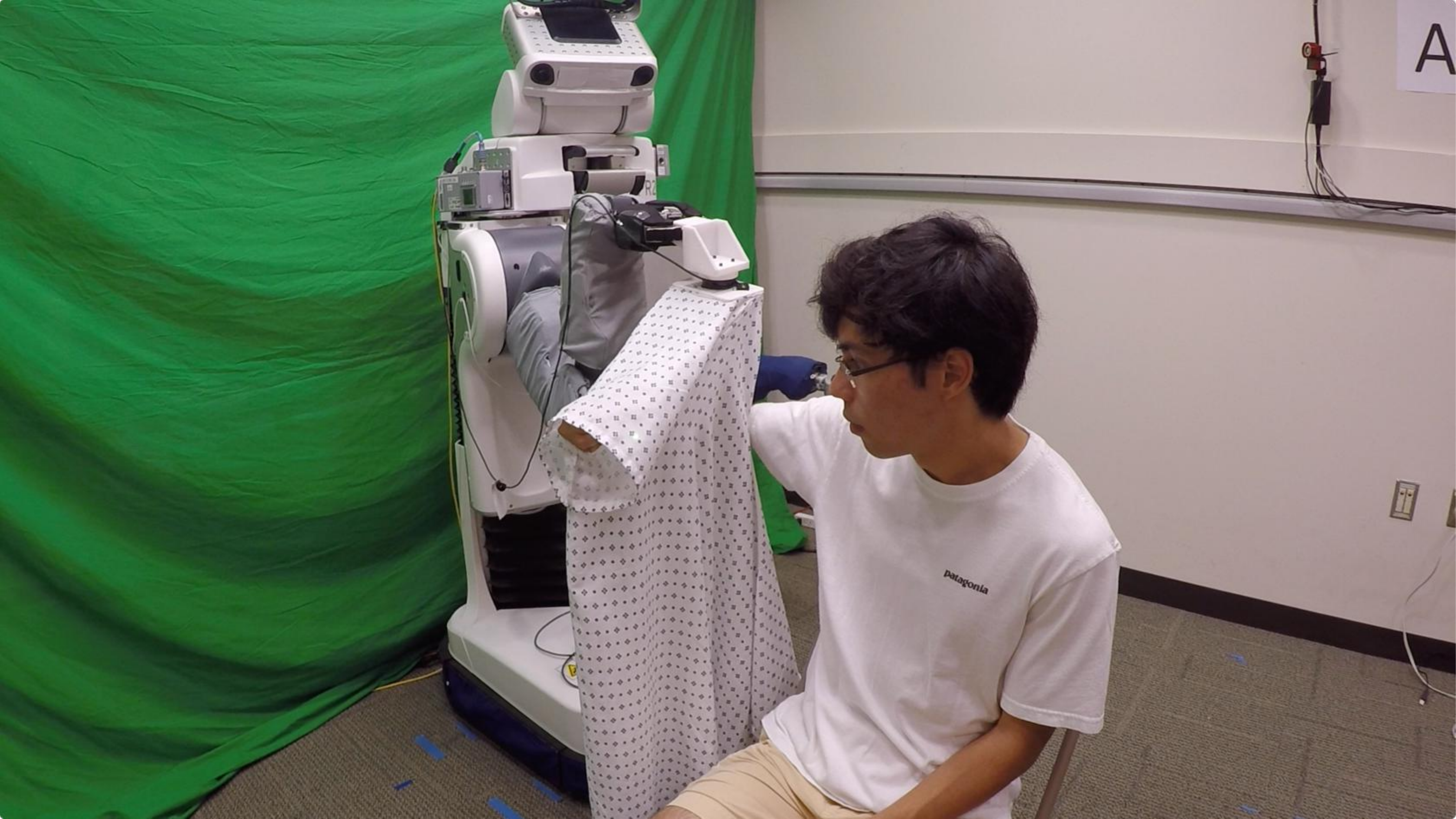}
\vspace{-0.4cm}
\caption{\label{fig:singlecatch}The gown getting caught on a participant's fist when the robot uses our MPC approach with a short prediction horizon of 0.01s.}
\vspace{-0.4cm}
\end{figure}

We evaluated our model predictive control approach on two robot-assisted dressing scenarios that involve pulling a hospital gown onto a participant's arm: (1) Full arm dressing: the robot must rotate its end effector to navigate around a participant's elbow and pull the garment onto the person's shoulder, as shown in Fig.~\ref{fig:intro}. (2) Circumvent a catch: the robot must predict that the garment will soon get caught on a person's fist, as seen in Fig.~\ref{fig:singlecatch}, and lower its end effector to avoid the catch. The robot performed 24 dressing trials per scenario, for a total of 48 trials per participant. We randomized the dressing scenarios and prediction horizons across all 48 trials. We updated the robot's actions at 5 Hz via Equation (\ref{eq:update}). We selected $w_1=0.5, w_2=20,$ and $w_3=0.5$ for our cost function presented in Equation (\ref{eq:cost}) as this empirically provided a balance between making task progress and keeping applied forces low.

For each scenario, we tested our method using three different prediction horizons, with 8 trials per horizon: 0.01s, 0.05s, and 0.2s. By testing multiple horizons, we show that a robot can better perform assistive tasks when it can predict the physical implications of its own actions. Note that changing the prediction horizon does not require model retraining since our model is capable of recursively predicting further into the future, which we discussed in Section~\ref{sec:simmodel}.

We used a Willow Garage PR2 robot to dress participants. The robot performed actions using the Orocos Kinematics and Dynamics Library\footnote{Orocos KDL: \url{http://www.orocos.org/kdl}}, which provided joint-level input to the PR2's low-level PID controllers. For participant safety, the PR2's arms were compliant and we set low PID gains for all arm joints. We zero out all forces and torques on the ATI force/torque sensor prior to a trial to account for the garment's weight. Additionally, we ran a force threshold monitor that halted all robot movement if forces measured at the robot's end effector exceeded 10~N. All computations to predict force maps for MPC were performed in real time on the robot's on-board CPUs.

Participants sat on a conventional folding chair and we instructed them to hold a specified static posture during each trial, shown in Fig.~\ref{fig:armpose}, and described below:

\begin{itemize}
\item Right arm bent 90 degrees at the elbow
\item Upper arm and forearm parallel with the ground
\item Fingers curled into a fist, knuckles vertically aligned
\item Thumb folded inwards over the fingers
\end{itemize}

\begin{figure*}
\centering
\includegraphics[width=0.19\textwidth, trim={15cm 8cm 14cm 1cm}, clip]{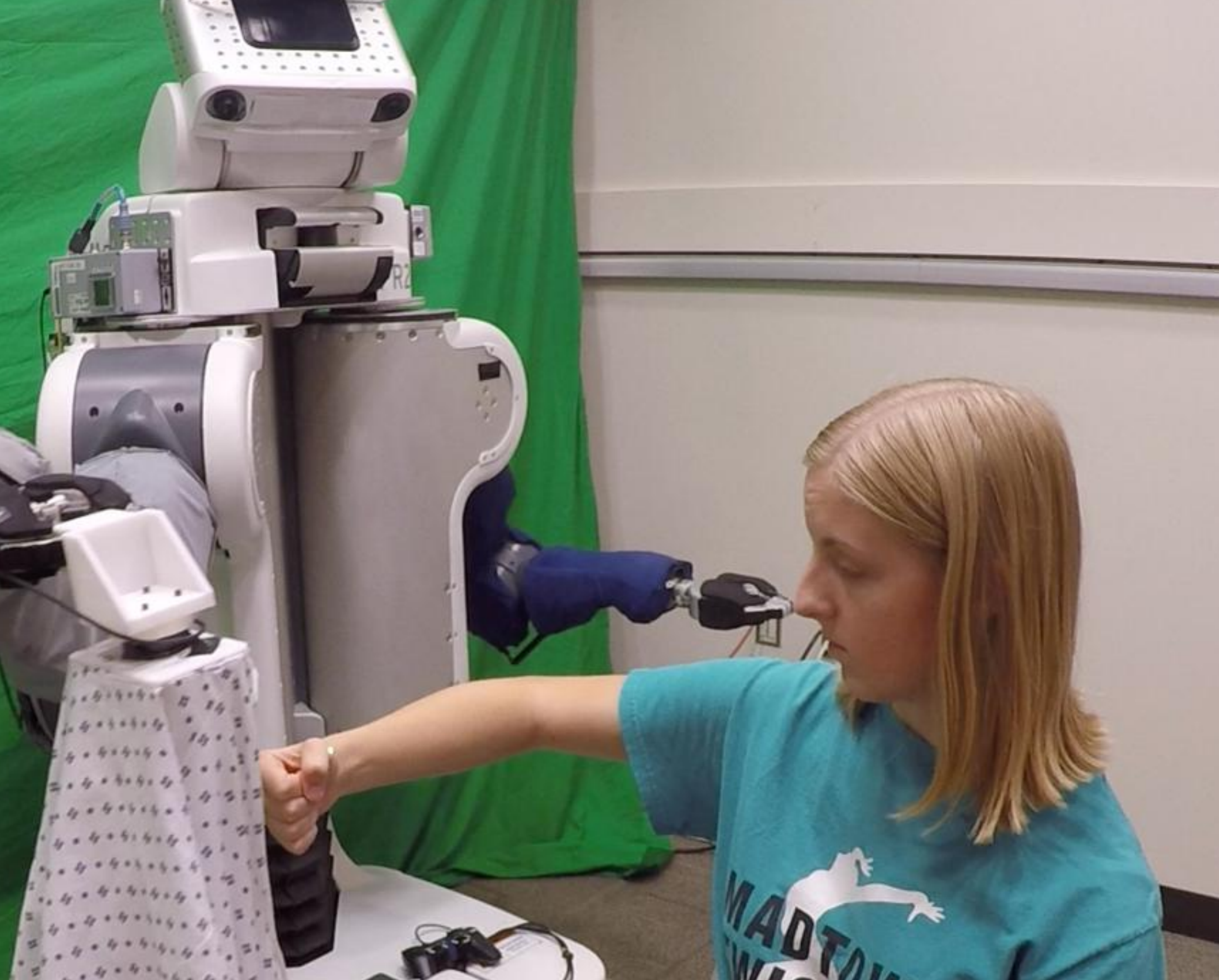}
\includegraphics[width=0.19\textwidth, trim={15cm 8cm 14cm 1cm}, clip]{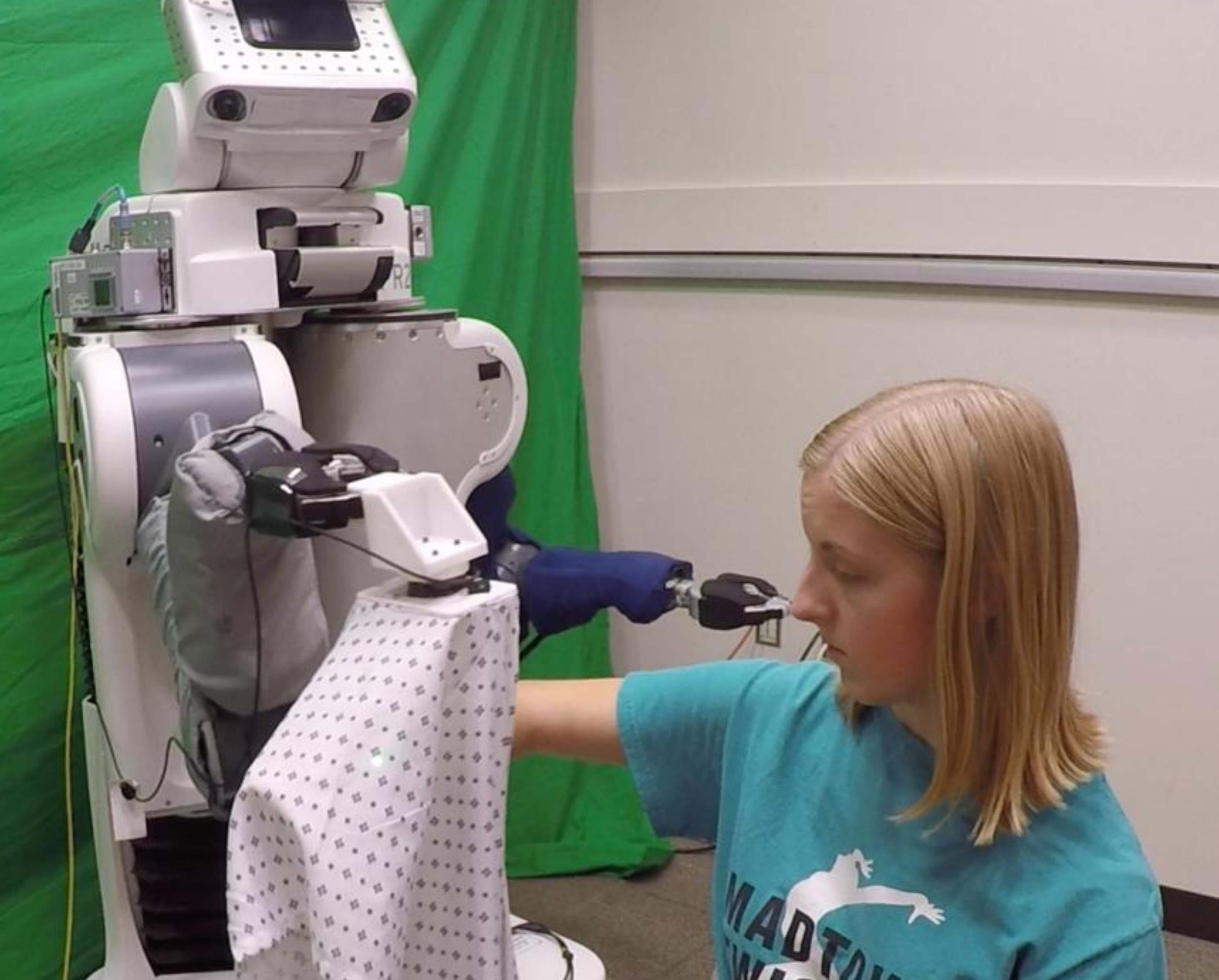}
\includegraphics[width=0.19\textwidth, trim={15cm 8cm 14cm 1cm}, clip]{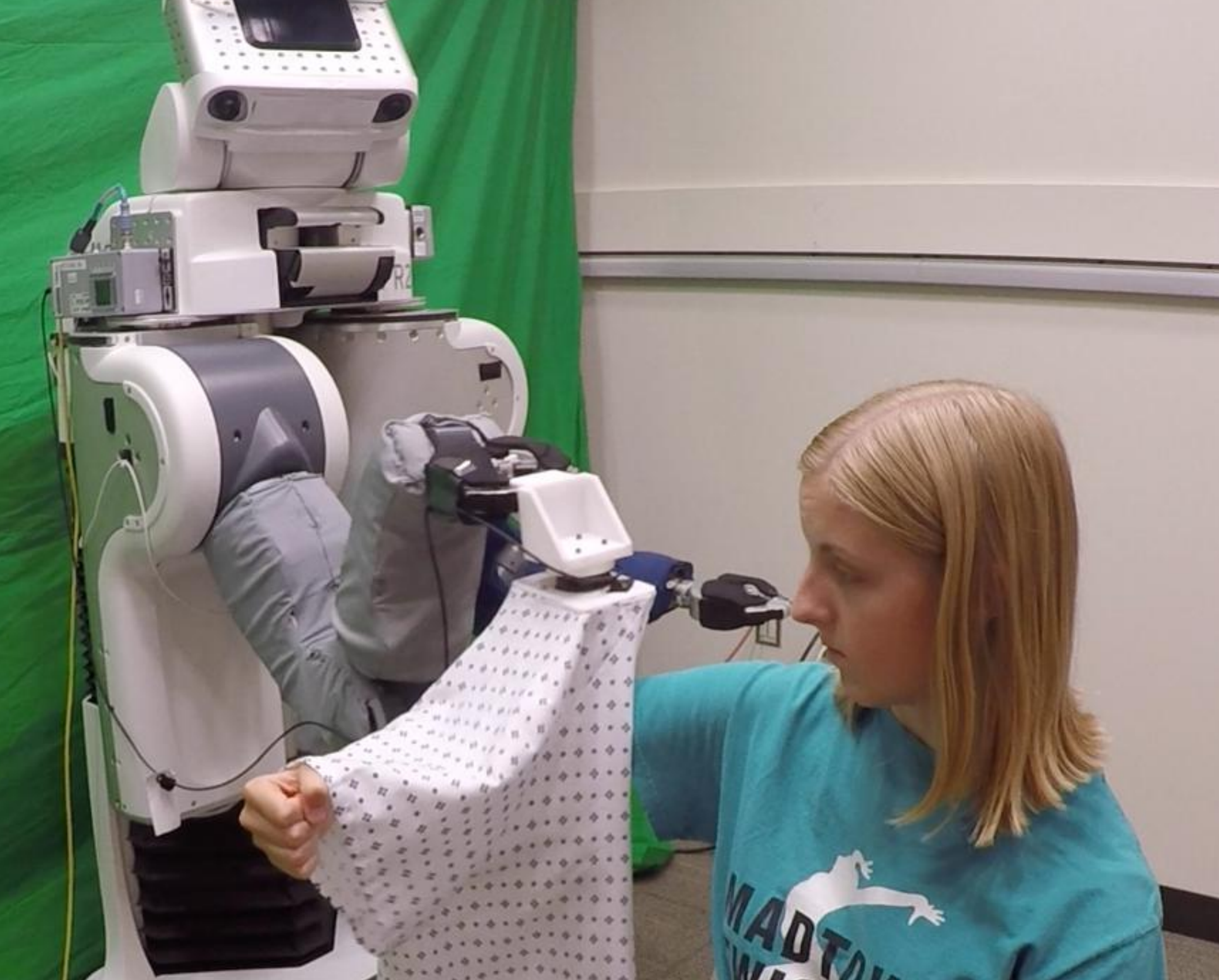}
\includegraphics[width=0.19\textwidth, trim={15cm 8cm 14cm 1cm}, clip]{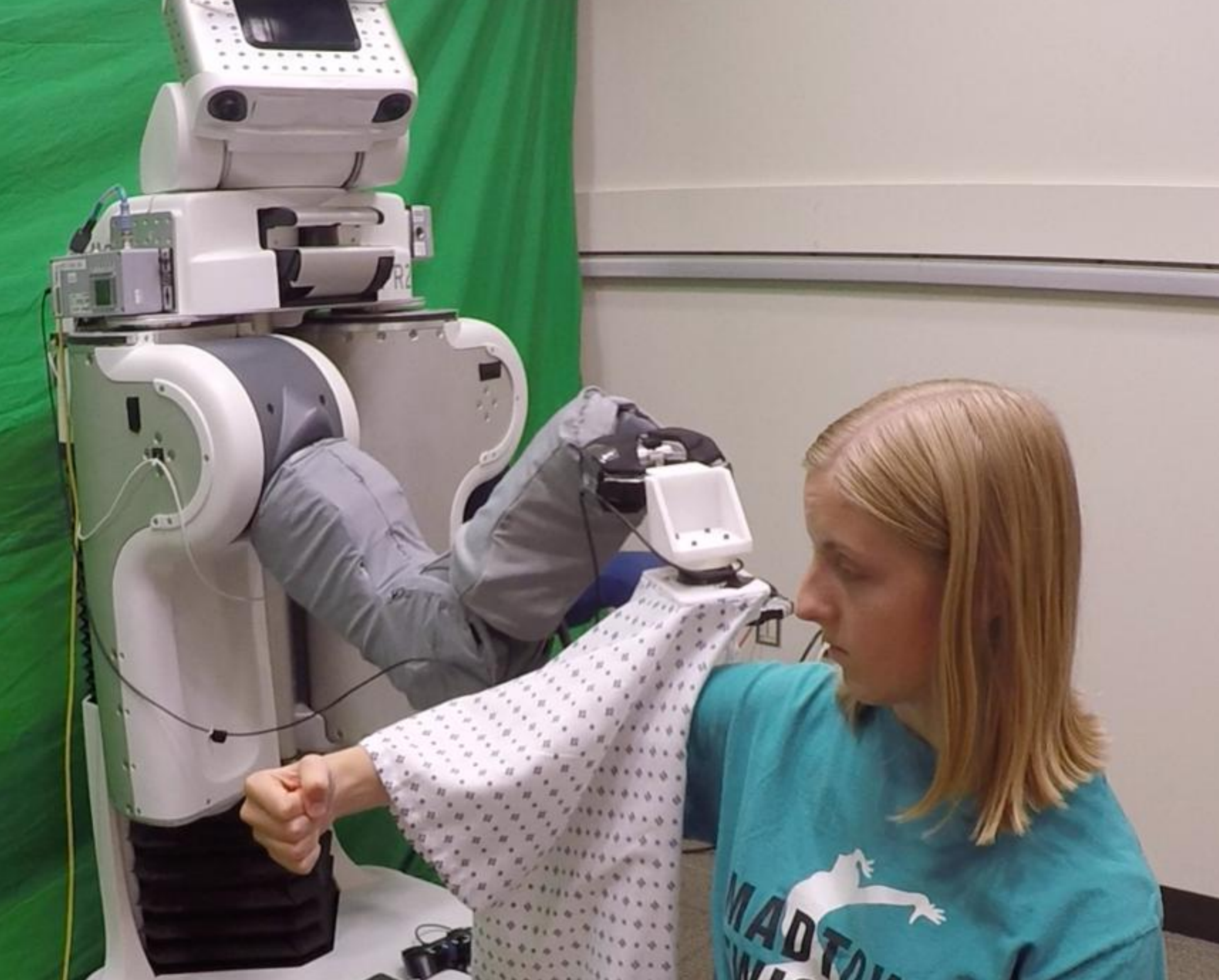}
\includegraphics[width=0.19\textwidth, trim={15cm 8cm 14cm 1cm}, clip]{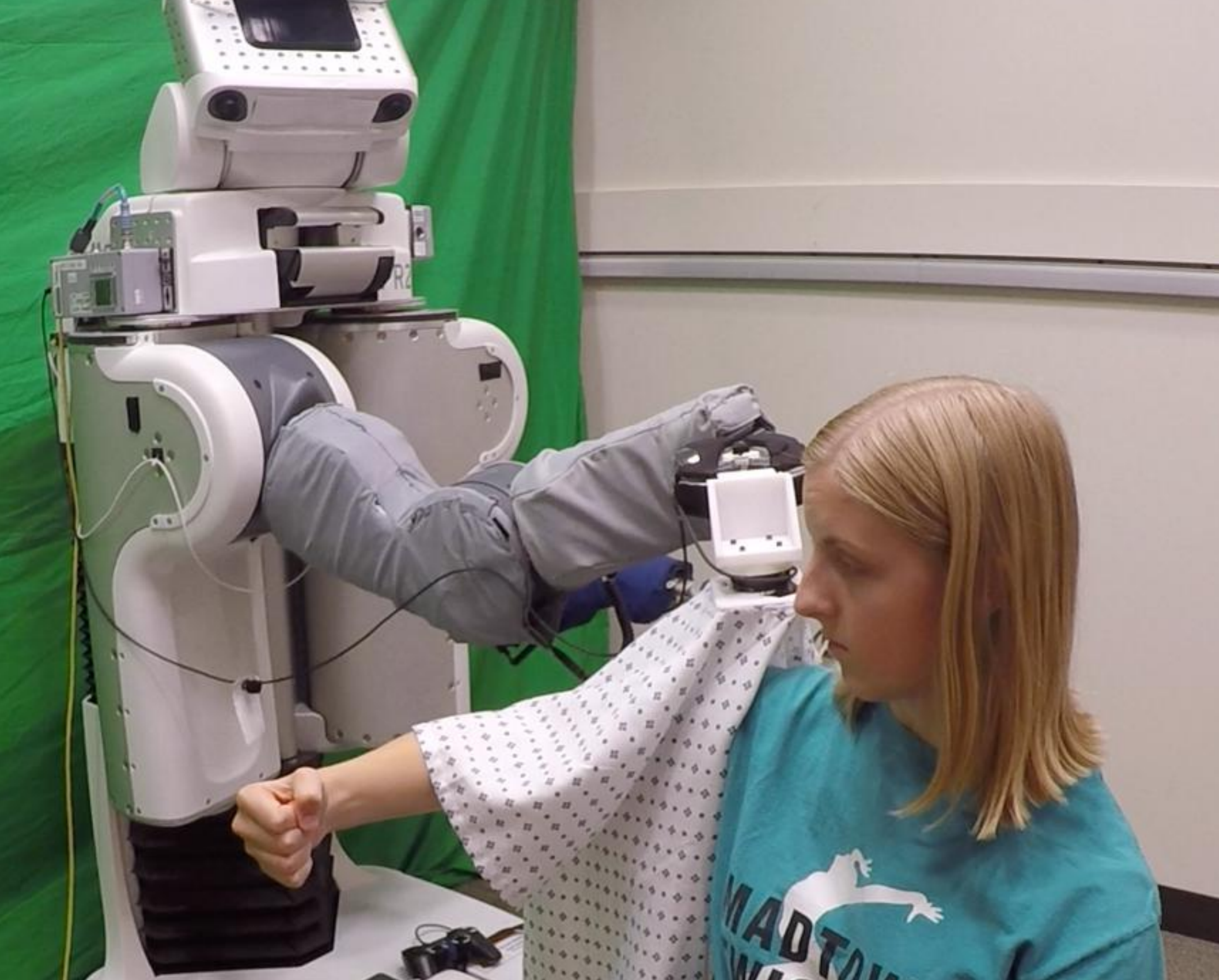}
\vspace{-0.1cm}
\caption{\label{fig:elbow}(Full Arm Dressing) With our predictive control approach and a horizon of 0.2s, the robot is able to navigate around a person's elbow and pull a gown up to a participant's shoulder.}
\end{figure*}

\begin{figure*}
\centering
\includegraphics[width=0.32\textwidth, trim={12cm 8cm 7cm 1cm}, clip]{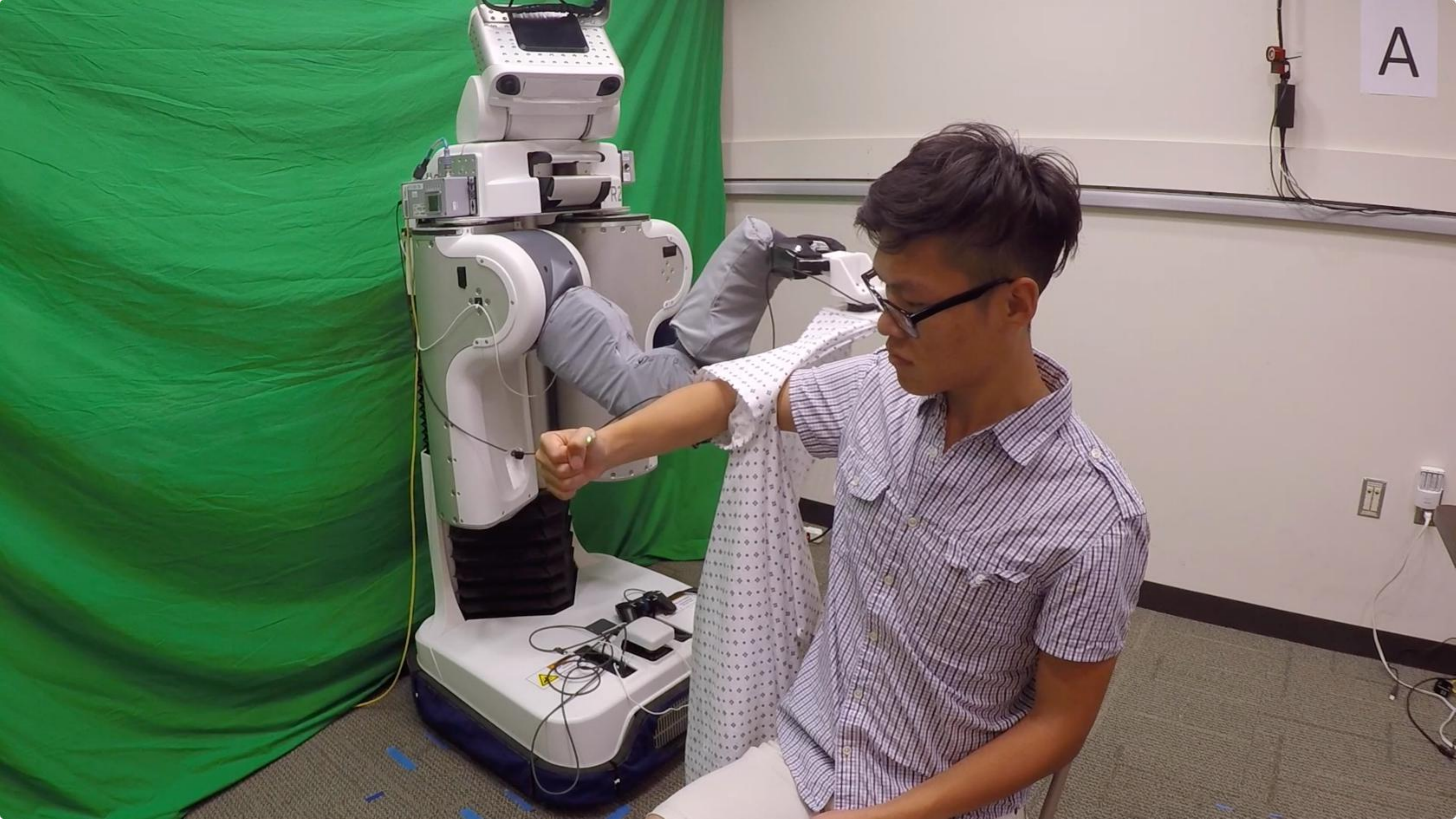}
\includegraphics[width=0.32\textwidth, trim={13cm 8cm 6cm 1cm}, clip]{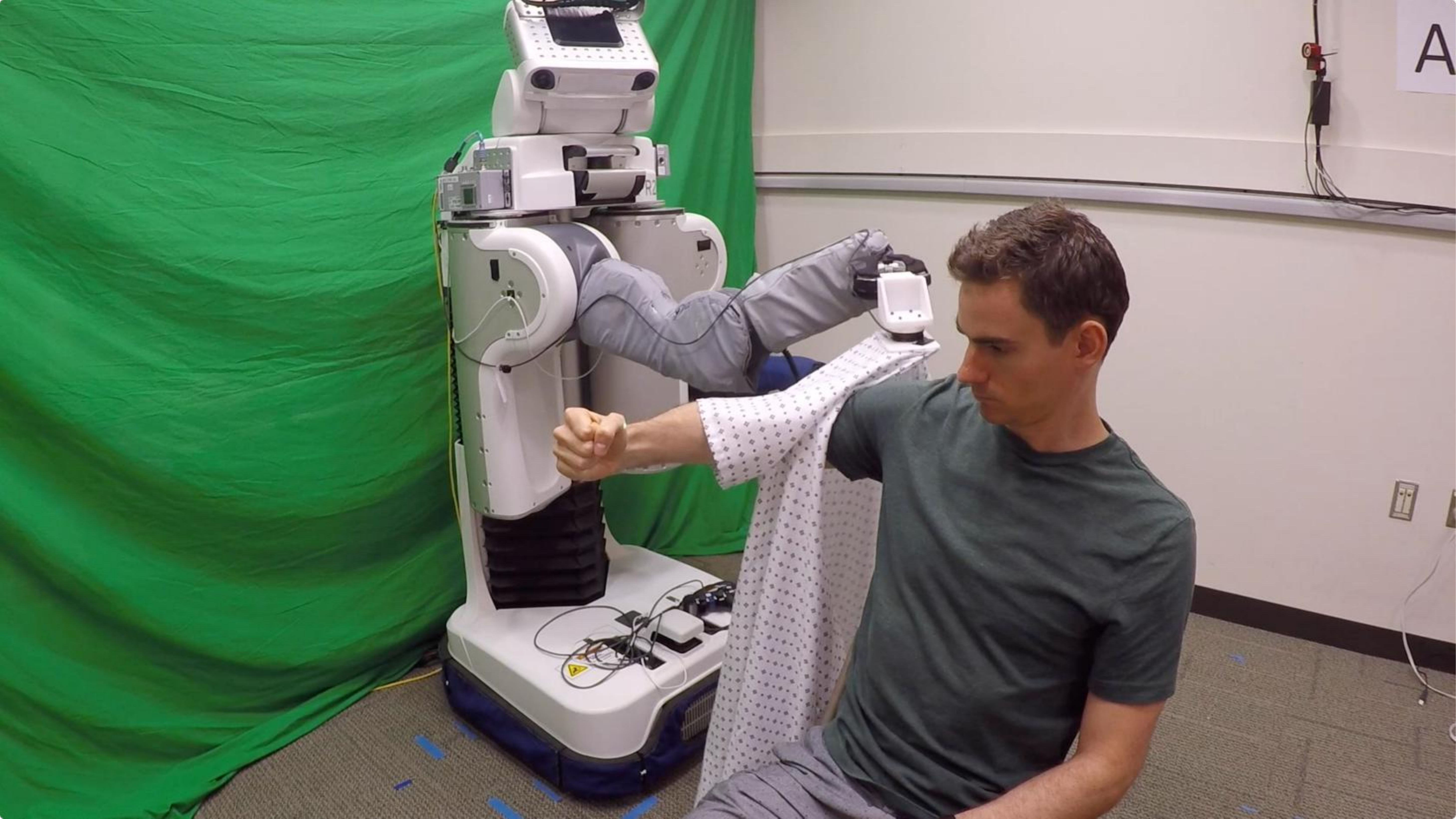}
\includegraphics[width=0.32\textwidth, trim={12cm 8cm 7cm 1cm}, clip]{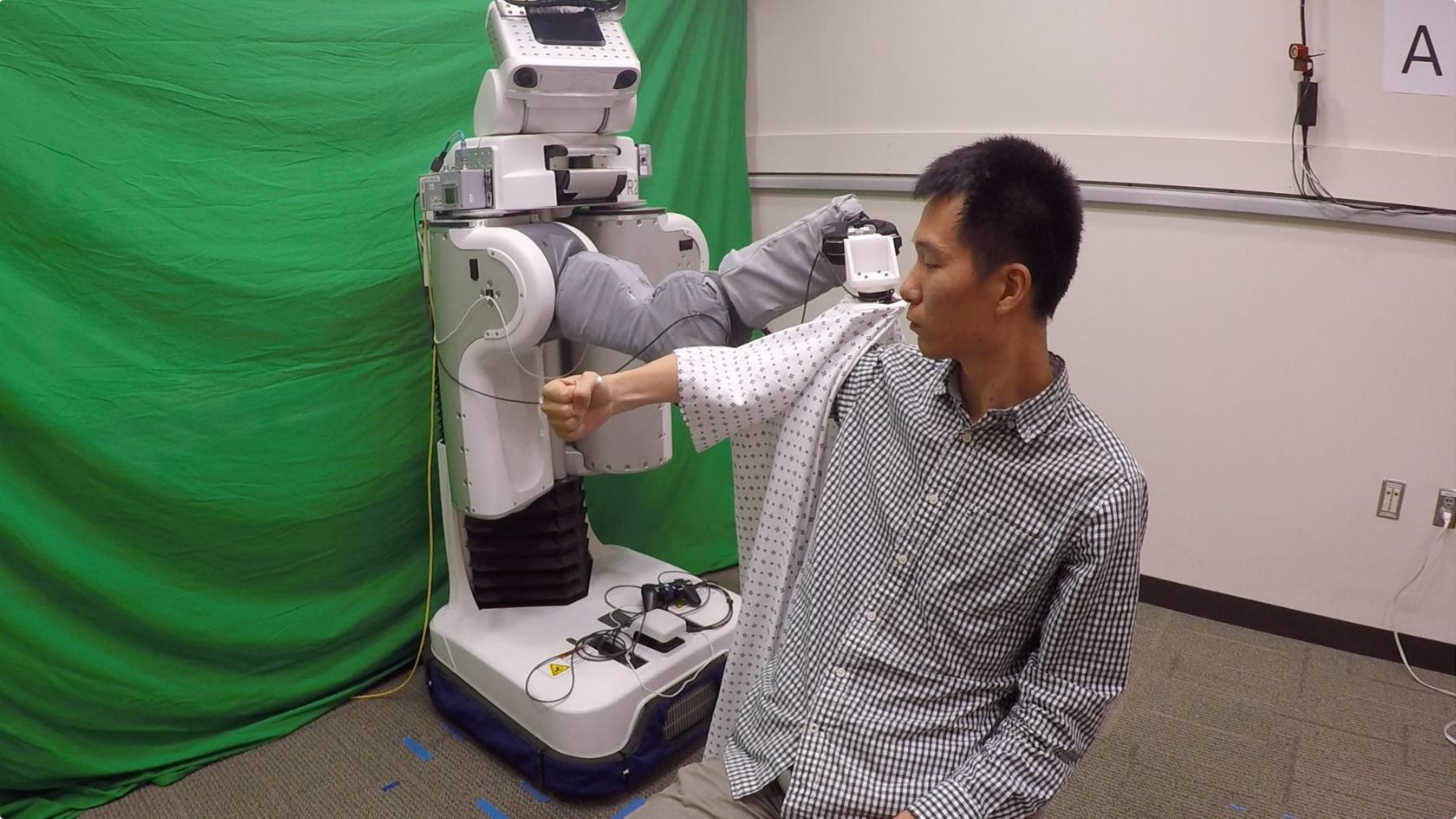}
\begin{tabularx}{\textwidth}{YYY}
    0.01s & 0.05s & 0.2s
\end{tabularx}
\vspace{-0.4cm}
\caption{\label{fig:elbow_endstate}(Full Arm Dressing) Example dressing outcomes for each of the three time horizons. The garment gets caught at a person's elbow for a time horizon of 0.01s, but our method successfully rotates the end effector and pulls the gown along a person's upper arm for horizons of 0.05s and 0.2s.}
\vspace{-0.4cm}
\end{figure*}

We set the initial robot configuration to hold the gown 15 cm in front of the participant's fist with the forearm direction normal to the opening in the gown. All participants started each trial seated comfortably while holding his or her arm in the specified posture. To promote consistency of arm position for appropriately comparing results with different prediction horizons, we used a commercial grade FDA approved laser pointer that pointed at the desired location for the participant's metacarpal-phalangeal joint---the base of the participant's thumb. We placed the laser on an adjustable height table to the left of the participant, facing the robot and orthogonal to the person's forearm, and we aligned the laser according to the participant’s height and posture.

We evaluated this work with participants who held a fixed arm pose, yet it may be preferable for a participant to hold their arm in different poses. We note that predicting the future forces applied to a person at varying poses remains an open problem and a limitation of our current work. Allowing the robot to estimate a person's pose prior to dressing, as seen in other works~\cite{gao2016iterative, yamazaki2014bottom}, may help alleviate this issue.

\begin{table}
\centering
\caption{\label{table:results}Task success averaged over all dressing assistance trials.}
\begin{tabular}{cccc} \toprule 
    & \multicolumn{3}{c}{Prediction Horizon} \\ \cmidrule{2-4}
    & 0.01s & 0.05s & 0.2s \\ \midrule\midrule
    \addlinespace[0.5em]
    Full Arm Dressing & 1.25\% & 97.5\% & 98.75\% \\
    \addlinespace[0.5em]
    Circumvent a Catch & 6.25\% & 26.25\% & 97.5\% \\
	\bottomrule
\end{tabular}
\vspace{-0.4cm}
\end{table}

\subsection{Full Arm Dressing}

For half of the dressing trials, we evaluated the robot's ability to navigate around the elbow and pull the garment entirely up the participant's arm. We were interested in what actions emerged when the controller's primary focus was to minimize the predicted forces applied to a participant's arm.

During a dressing trial, the robot selected actions that minimized the cost from Equation (\ref{eq:cost}). Each trial began with the PR2 holding the top of the gown opening 10~cm above the top of a participant's fist. We marked the end of a dressing trial whenever the magnitude of forces measured at the end effector exceeded 10 N, or the robot's arm reached its joint limits. For the full arm dressing trials, we classified a trial as successful if the trial completed without reaching the force threshold and the inner seam on the sleeve, defined by where the sleeve is sewn onto the main body of the gown, had passed the participant's elbow. Fig.~\ref{fig:elbow} shows a successful sequence of this dressing scenario when the robot used our MPC method with a prediction horizon of 0.2s (20 time steps). Note that once the robot's end effector reaches a person's elbow, the robot can continue to minimize applied forces by performing a yaw rotation to navigate around the elbow and begin moving along the upper arm. This results in the robot pulling the garment entirely up a person's arm.

\begin{figure}
\centering
\includegraphics[width=0.48\textwidth, trim={1cm 18.0cm 2cm 3cm}, clip]{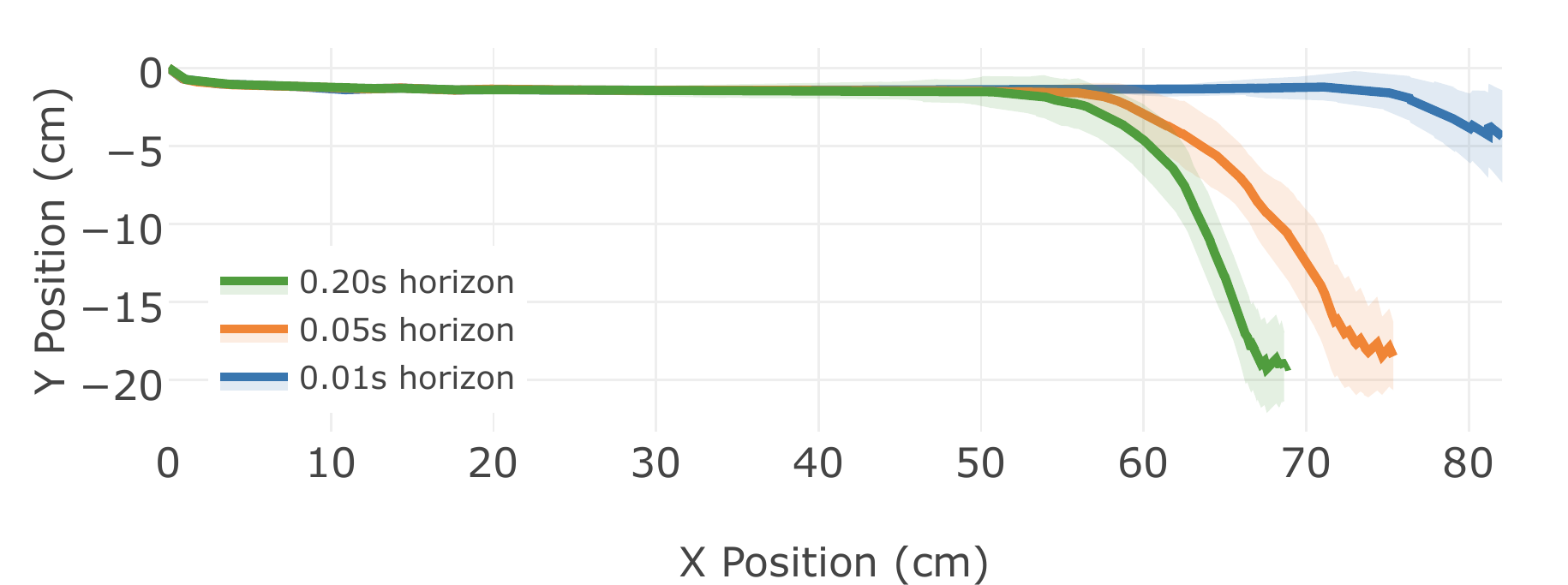}
\vspace{-0.4cm}
\caption{\label{fig:elbowposition}Top-down view of the end effector's path for the three prediction horizons during the full arm dressing trials. Paths are averaged across all 10 participants and background shading represents one standard deviation. The robot successfully dressed a person's arm with a 0.05s and 0.2s prediction horizon, yet the robot failed to rotate around the elbow with a 0.01s horizon.}
\vspace{-0.4cm}
\end{figure}

\begin{figure}
\centering
\includegraphics[width=0.48\textwidth, trim={1cm 18.0cm 2cm 3cm}, clip]{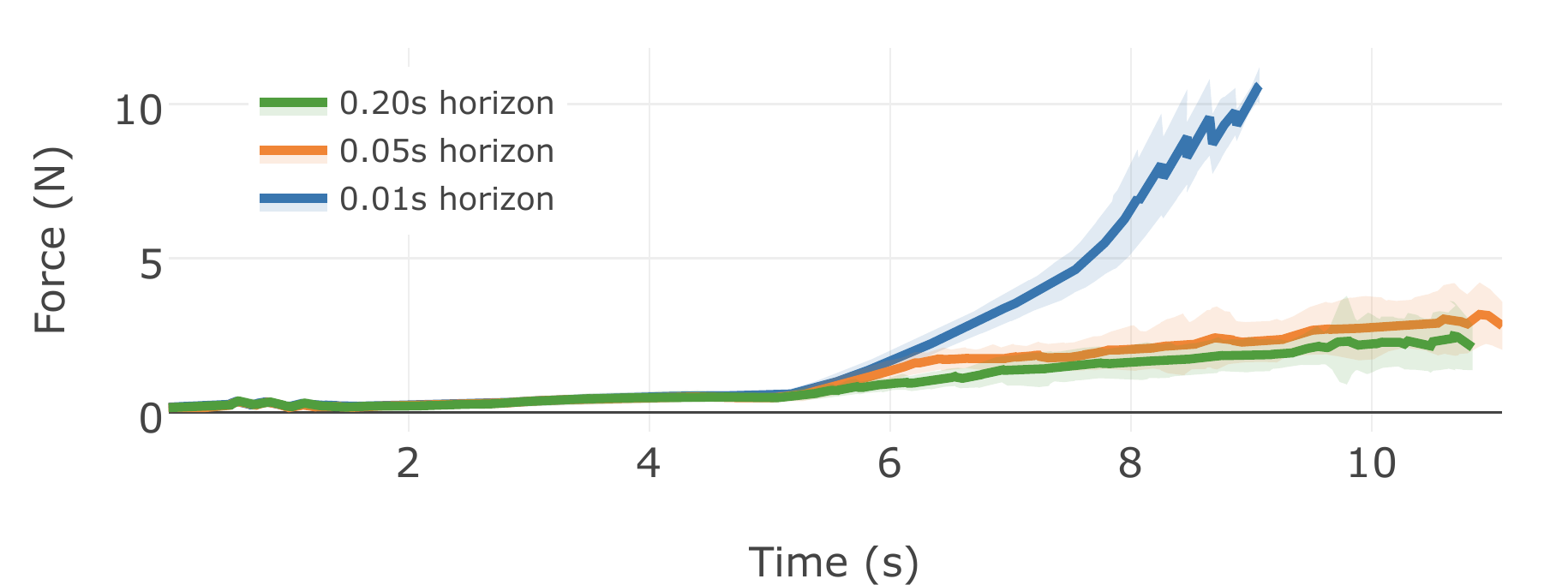}
\vspace{-0.4cm}
\caption{\label{fig:elbowforce}Magnitude of force measured by the ATI force/torque sensor at the robot's end effector during the full arm dressing trials. Results are averaged across all 10 participants and background shading represents one standard deviation. The run time for each horizon differs based on when the force threshold or joint limits were reached. With a shorter prediction horizon of 0.01s, the controller applied undesirable high forces at a person's elbow.}
\vspace{-0.4cm}
\end{figure}

\begin{figure*}
\centering
\includegraphics[width=0.19\textwidth, trim={18cm 10cm 18cm 8cm}, clip]{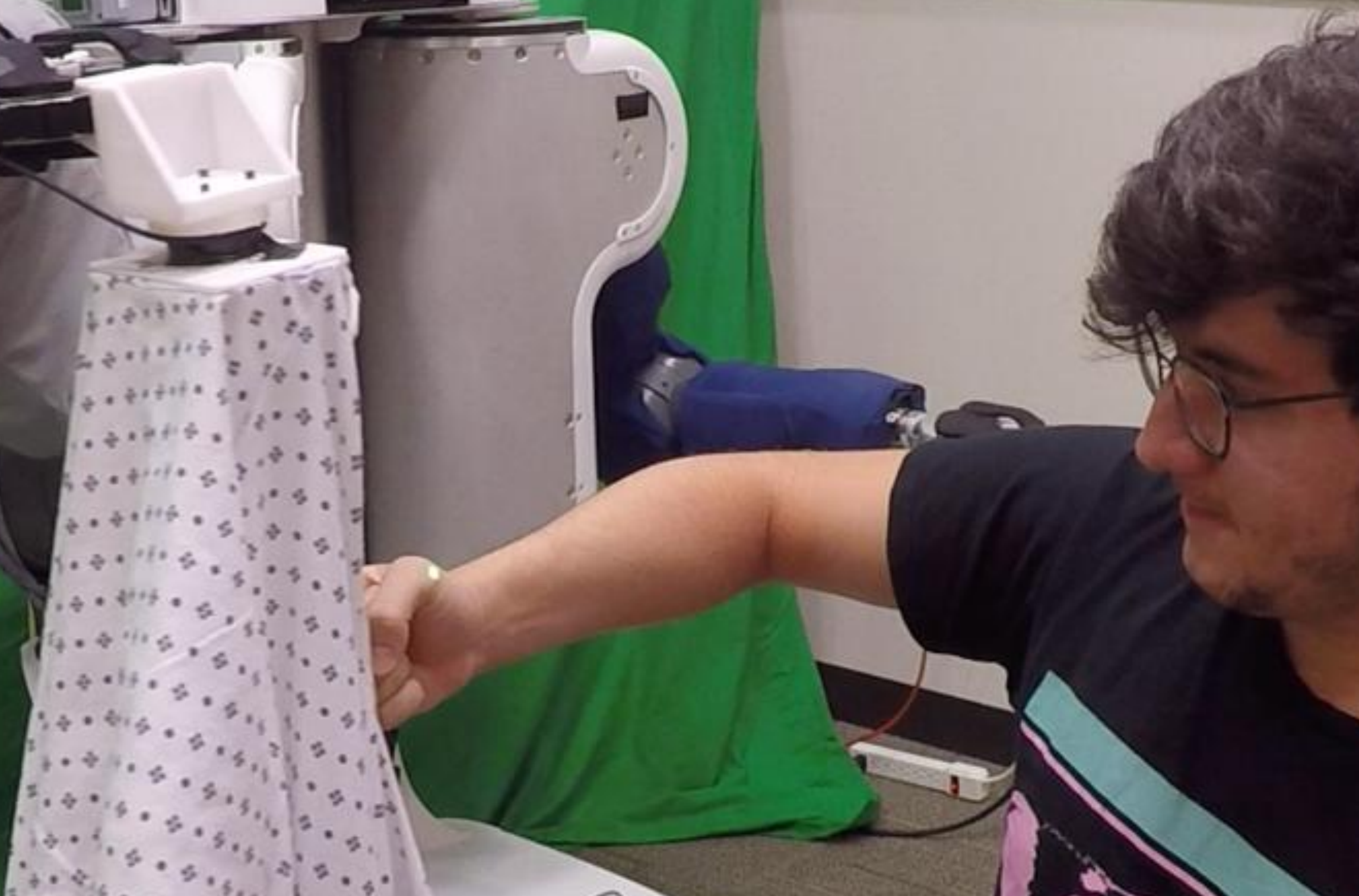}
\includegraphics[width=0.19\textwidth, trim={18cm 10cm 18cm 8cm}, clip]{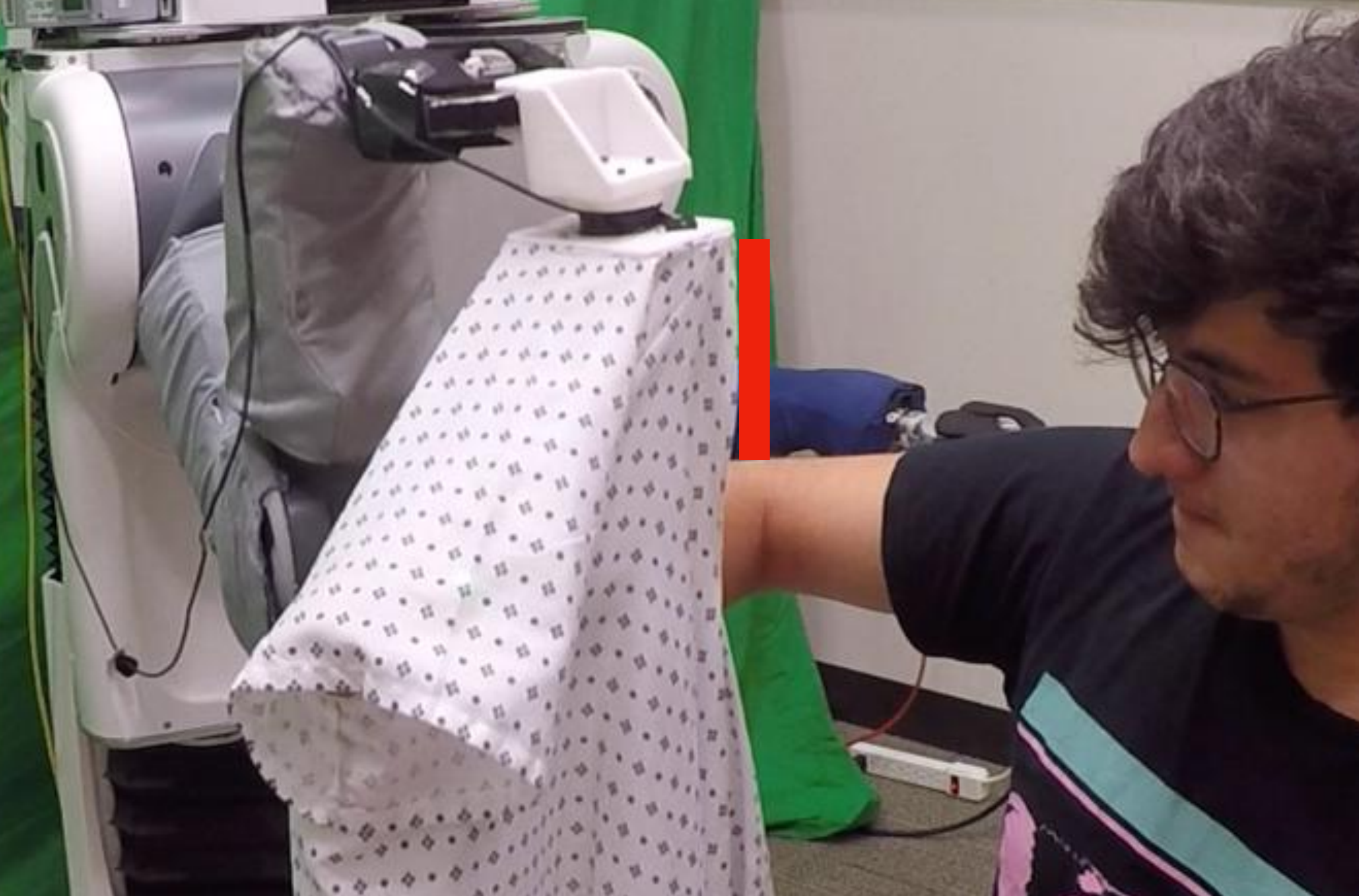}
\includegraphics[width=0.19\textwidth, trim={18cm 10cm 18cm 8cm}, clip]{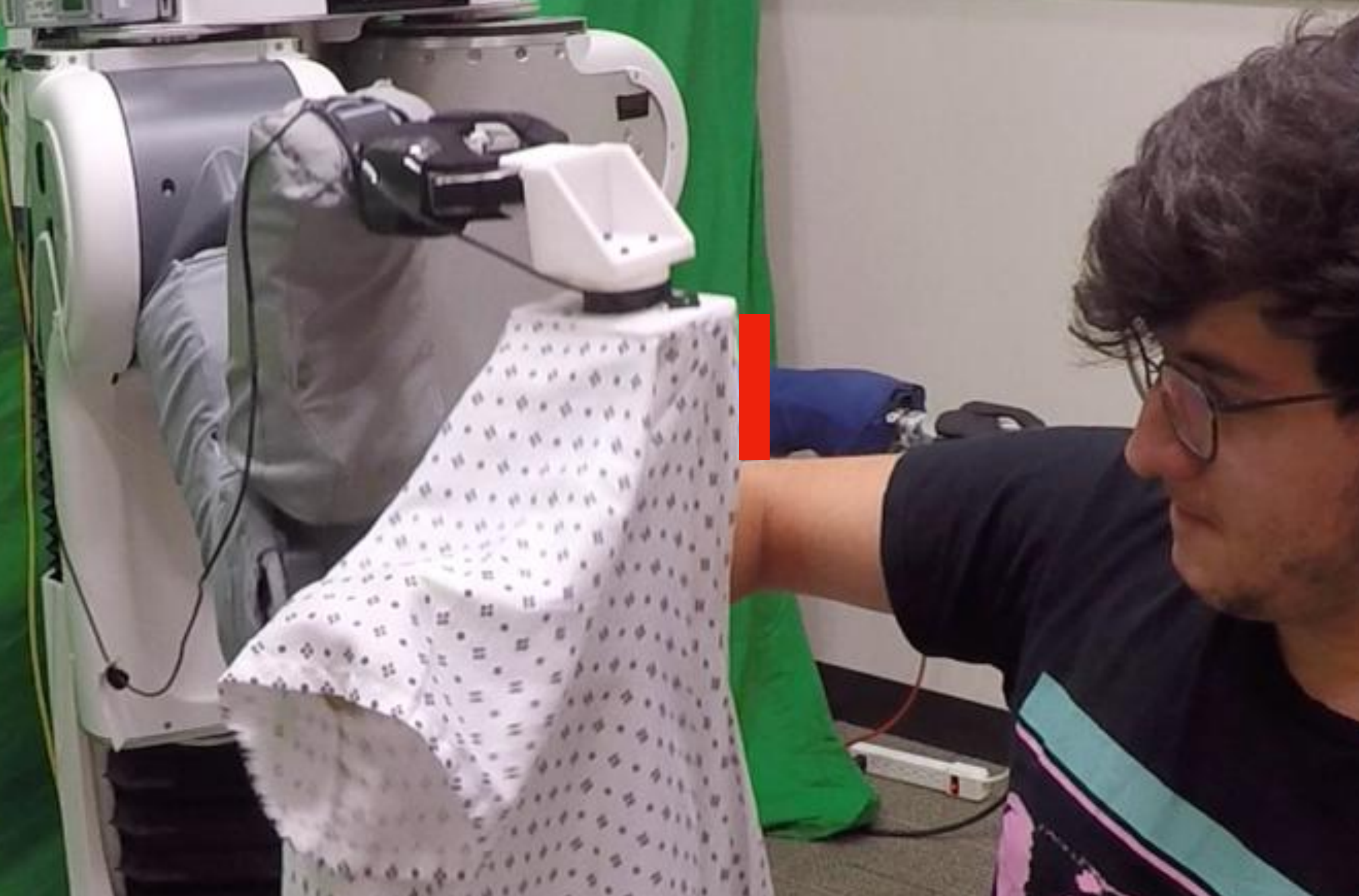}
\includegraphics[width=0.19\textwidth, trim={18cm 10cm 18cm 8cm}, clip]{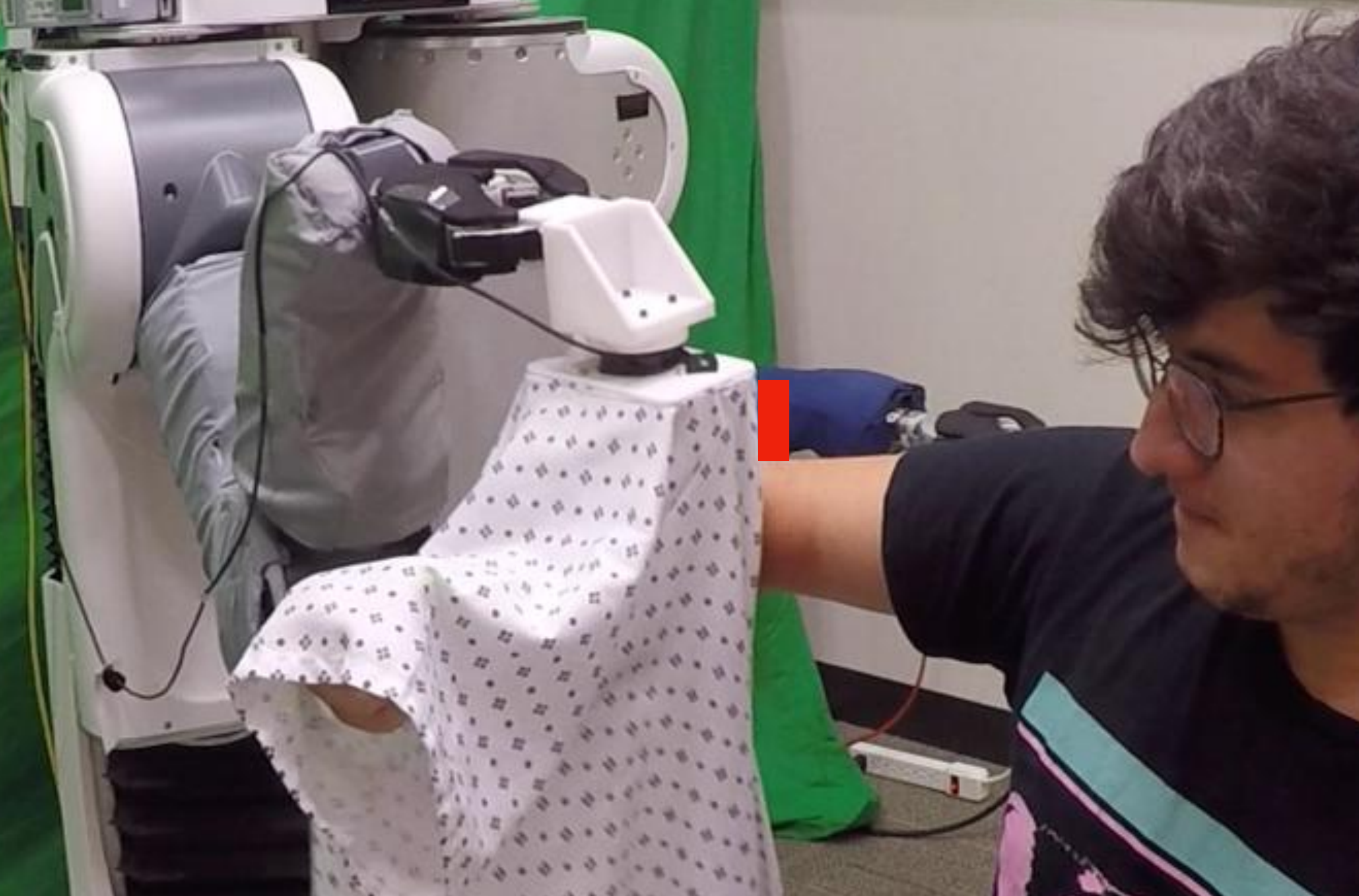}
\includegraphics[width=0.19\textwidth, trim={18cm 10cm 18cm 8cm}, clip]{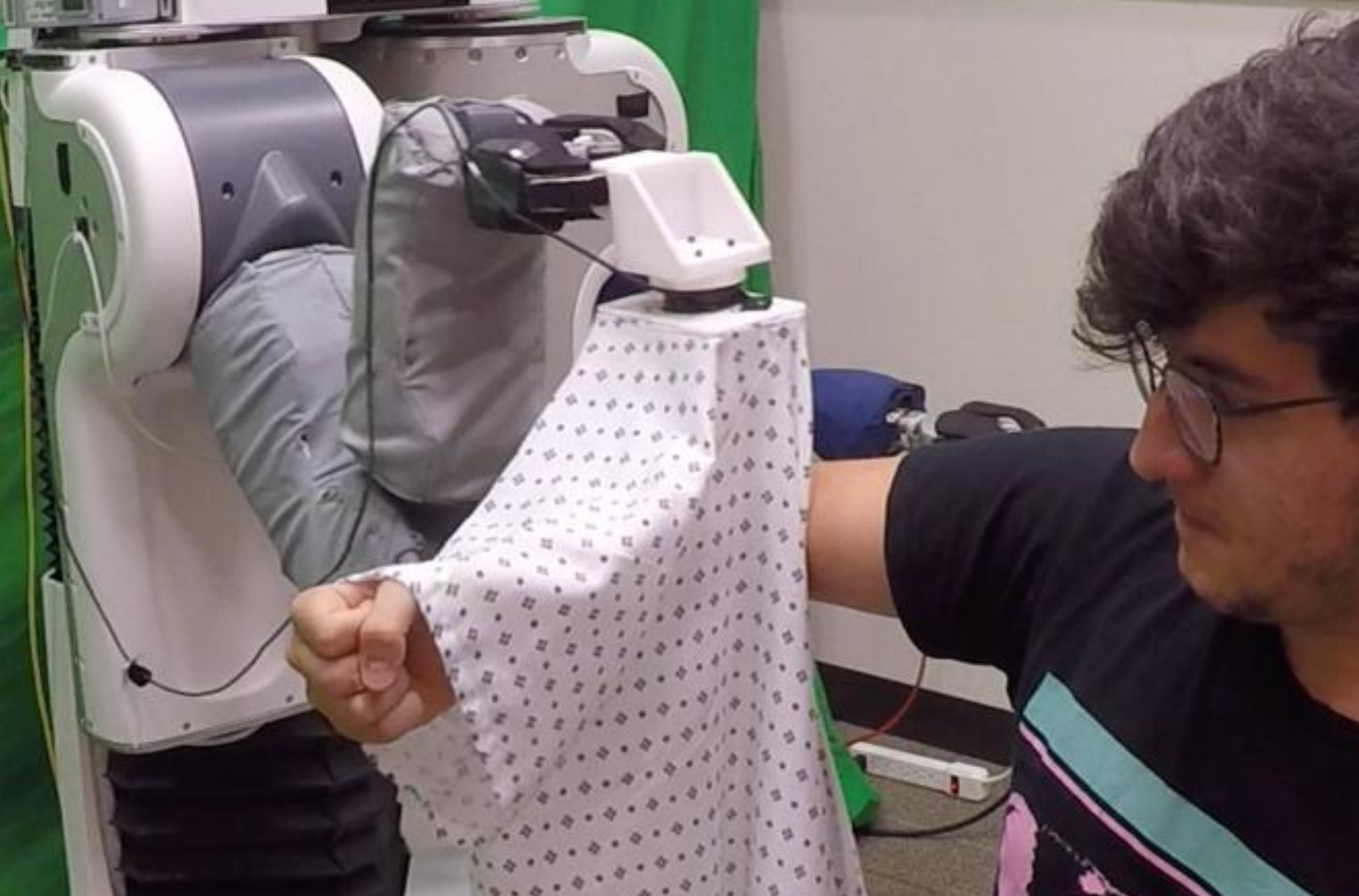}
\vspace{-0.1cm}
\caption{\label{fig:catch}(Circumventing a Catch) With a horizon of 0.2s, our approach predicts that the garment will soon get caught on a participant's fist and apply large forces onto the fist. Note how the robot's end effector drops closer to the participant's arm, shown by the red line, to avoid the garment from catching.}
\end{figure*}

\begin{figure*}
\centering
\includegraphics[width=0.32\textwidth, trim={15cm 8cm 8cm 1cm}, clip]{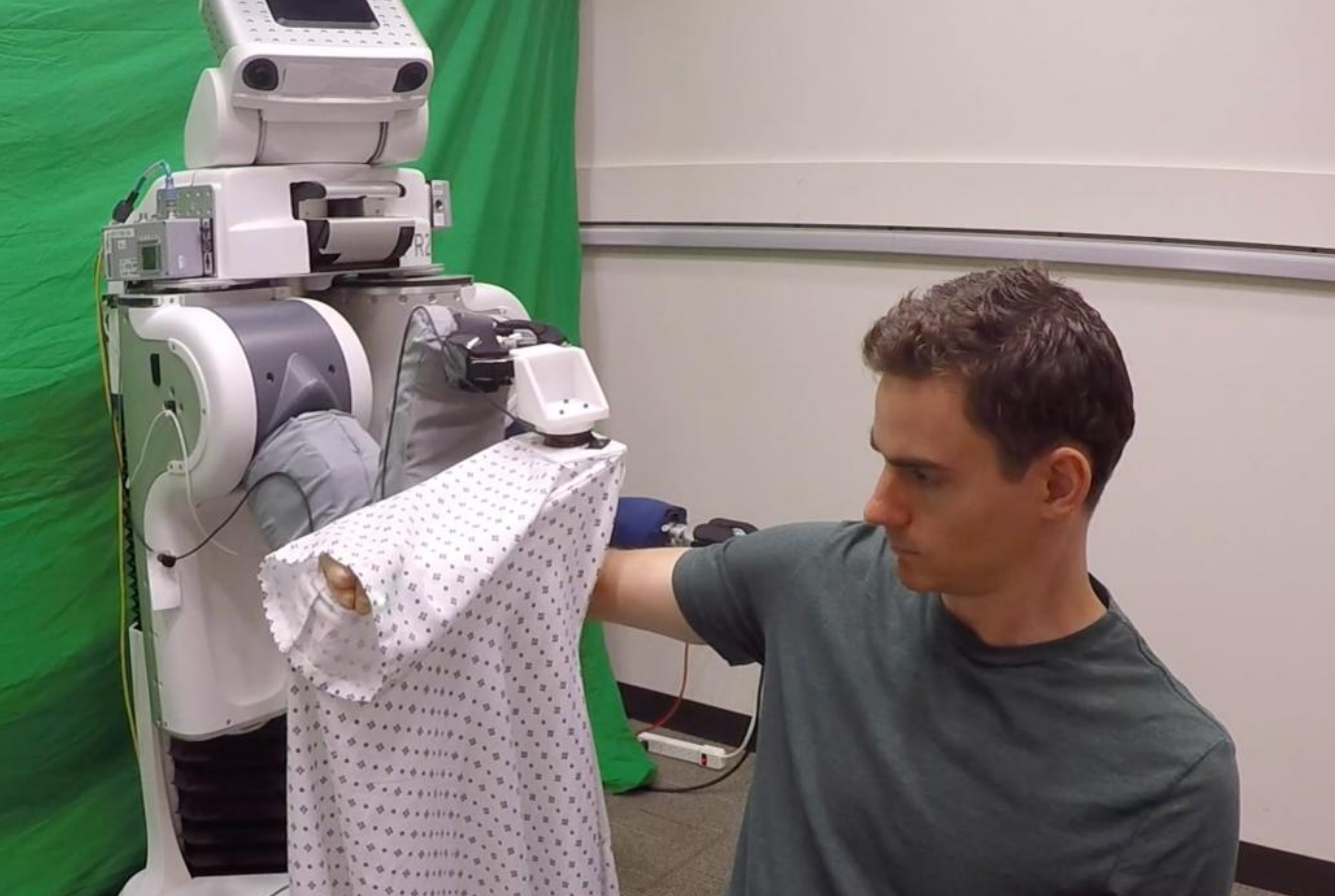}
\includegraphics[width=0.32\textwidth, trim={14cm 7.5cm 9cm 1.5cm}, clip]{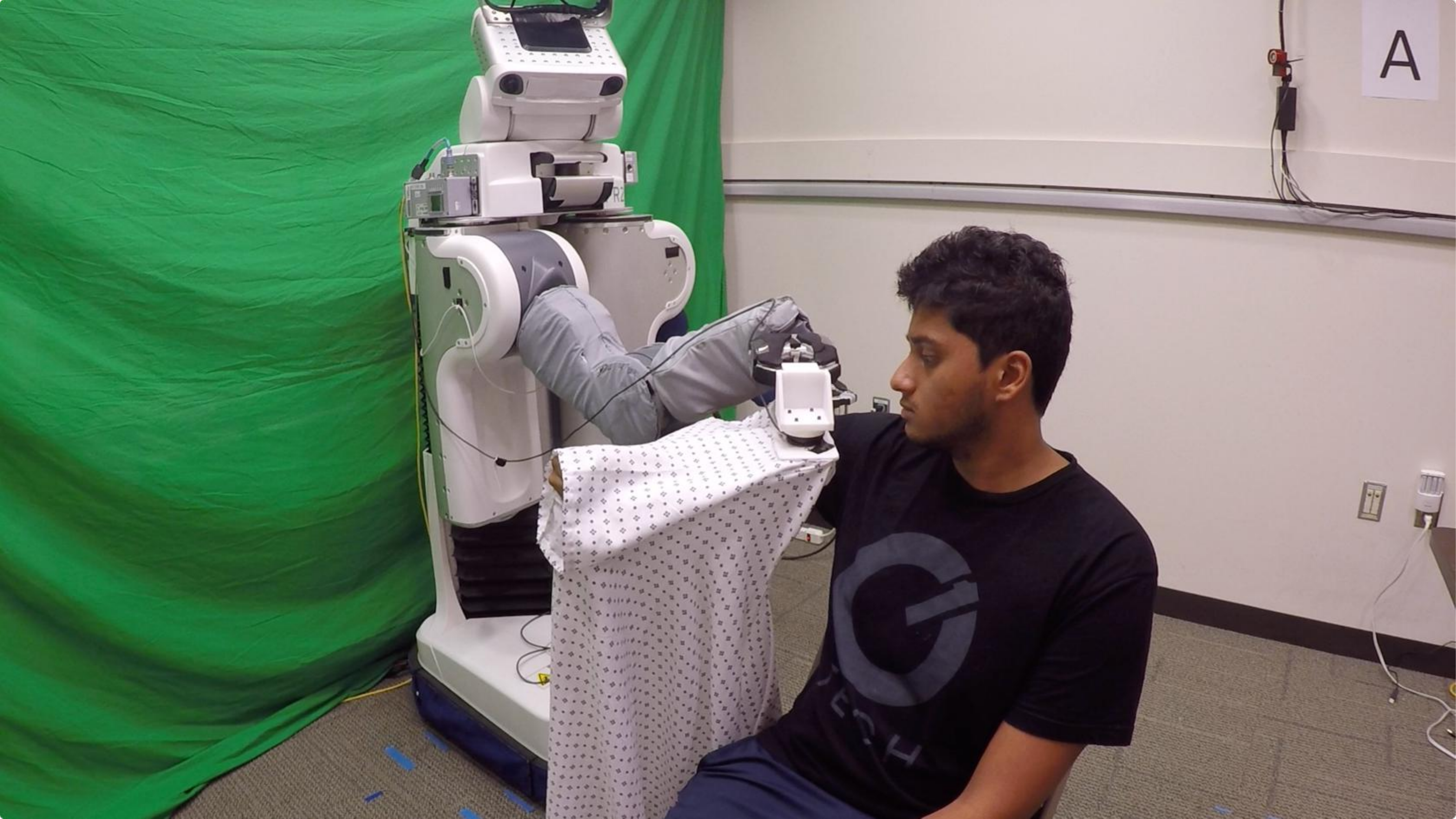}
\includegraphics[width=0.32\textwidth, trim={12cm 7.5cm 11cm 1.5cm}, clip]{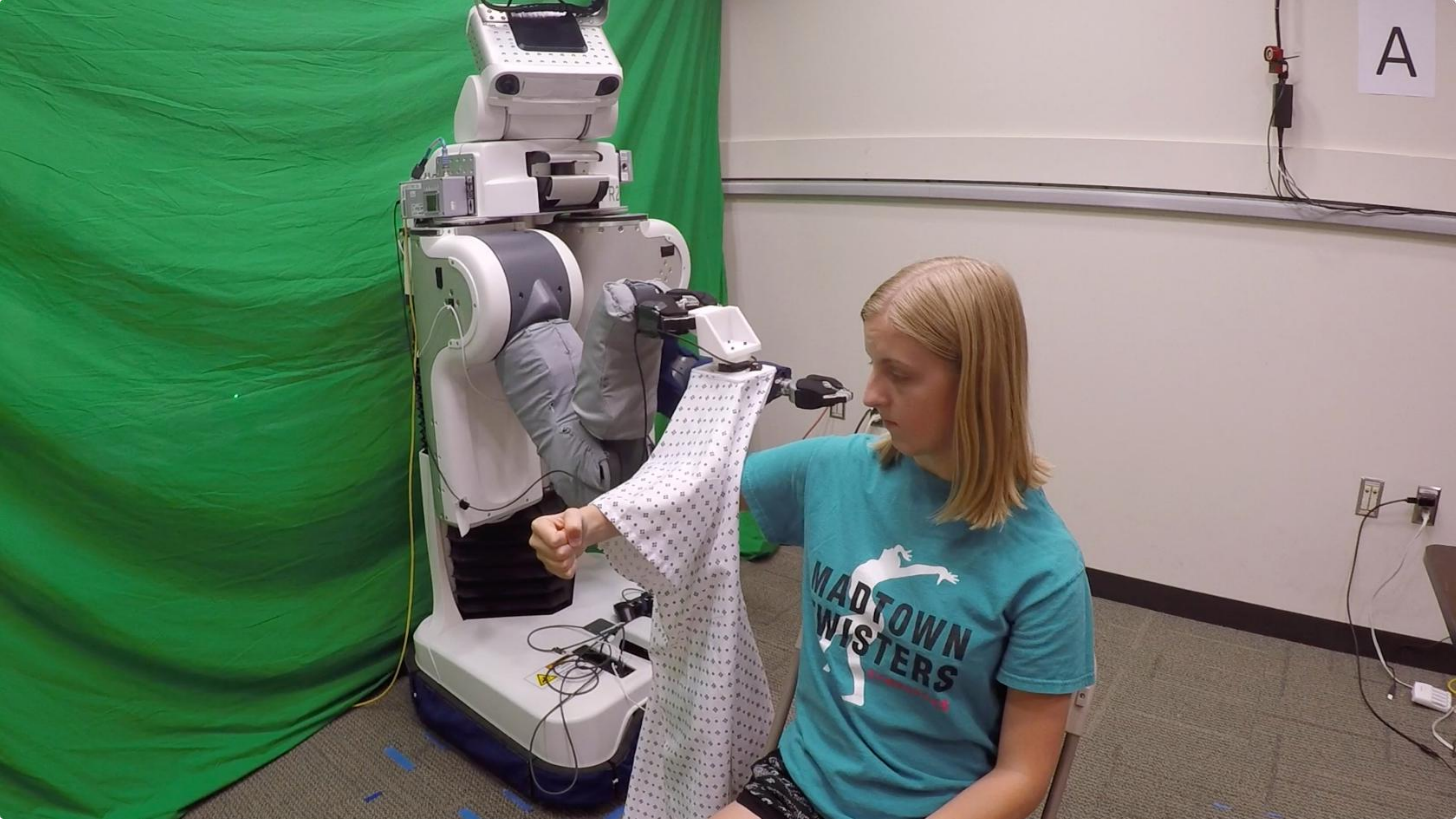}
\begin{tabularx}{\textwidth}{YYY}
    0.01s & 0.05s & 0.2s
\end{tabularx}
\vspace{-0.4cm}
\caption{\label{fig:catch_endstate}(Circumventing a Catch) Example dressing outcomes when our controller attempts to avoid the garment from catching on a participant's fist. The garment often gets caught at a person's fist for horizons of 0.01s and 0.05s, yet our approach successfully navigates around the catch with a 0.2s horizon.}
\vspace{-0.4cm}
\end{figure*}

\begin{figure}
\centering
\includegraphics[width=0.48\textwidth, trim={1cm 18.0cm 2cm 3.5cm}, clip]{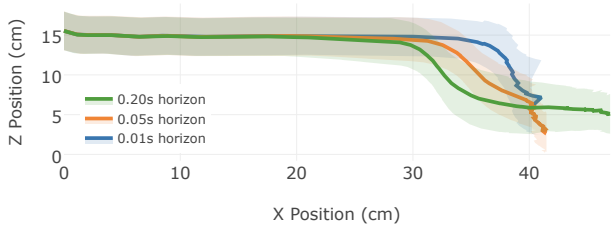}
\vspace{-0.4cm}
\caption{\label{fig:catchposition}Side view of the end effector's path for the three horizons while the robot attempted to circumvent a catch. Paths are averaged across trials from all 10 participants and background shading depicts one standard deviation.}
\vspace{-0.4cm}
\end{figure}

In Fig.~\ref{fig:elbow_endstate}, we display outcomes of dressing trials for the three prediction horizons. For a horizon of 0.01s, the predicted force maps across candidate actions are nearly identical. Because of this, the robot was unable to find an action that significantly lowered applied forces and instead continued to pull the garment into a person's elbow until the 10 N threshold was reached. In contrast, both the 0.05s and 0.2s horizons led to the robot rotating its end effector and pulling the garment up to a participant's shoulder, successfully navigating around the person's elbow. Both Fig.~\ref{fig:elbow} and the supplementary video show this procedure in detail. 

Fig.~\ref{fig:elbowposition} shows a top-down view of the end effector path for each prediction horizon, averaged across all 10 participants. A horizon of 0.2s led to the robot rotating and moving along the upper arm sooner than for a horizon of 0.05s, yet both led to actions that fully dressed a person's arm. The task success rates for each prediction horizon can be found in Table~\ref{table:results}. These success rates are averaged over 80 trials for each scenario. Fig.~\ref{fig:elbowforce} displays the magnitude of the force measured at the robot's end effector across trials for all 10 participants. For a 0.01s horizon, we again notice that the robot continues to apply more force on a person's elbow until it reaches the 10~N threshold.

When contact occurs between the garment and a person's body, our control approach can use haptic and kinematic observations to dress a person by primarily minimizing predicted forces. Yet, a limitation of this purely haptic and kinematic-based approach is that the controller is provided with no information about a person's initial pose. As a result, the robot would be unable to recognize or replan actions if the garment were to entirely miss a person's body. Future work could address this by incorporating other modalities, such as vision-based techniques, to estimate a person's pose before or during dressing~\cite{klee2015personalized,chance2016assistive}.

\subsection{Circumvent a Catch}

In this section, we evaluate our model's ability to predict that a garment will get caught and apply large force onto a participant's fist. During these trials, we also evaluated how well our MPC approach selected actions that properly averted the catch in order to reduce predicted forces. We adjusted the starting height of the robot's end effector according to each participant's arm height. Specifically, we aligned the end effector so that the bottom seam of the sleeve would get caught in the middle of a participant's fist when the robot followed a forward linear trajectory. A dressing trial ended whenever the end effector forces exceeded 10 N, or the end effector reached the participant's elbow, along the X-axis. A trial was successful if the end effector reached the elbow along the X-axis without exceeding the force threshold. 

Fig.~\ref{fig:catch} presents a sequences of images for a successful trial with a 0.2s horizon in which the robot's end effector would drop down closer to a participant's forearm to bypass the catch. Notice that the robot could also choose to lift its end effector to avoid the catch. The robot may not have chosen to lift up over the hand due to the forces that occur when the entire garment drags across a person's fist.

Fig.~\ref{fig:catch_endstate} shows example outcomes of dressing trials for each of the three prediction horizons. A horizon of 0.01s consistently led to the garment getting caught on a person's fist for 93.75\% of the trials, as shown in Table~\ref{table:results}. A prediction horizon of 0.05s also failed to avoid the catch for most trials.

Finally, Fig.~\ref{fig:catchposition} shows a side view of the end effector path for each prediction horizon, averaged across all 10 participants. As shown, the horizon length impacts how soon our approach detects the catch and replans. The controller attempted to move the end effector downwards to avert the catch for all three prediction horizons. However, timing is crucial and only the 0.2s horizon allowed our method to detect the catch soon enough to consistently avoid it.

Overall, these results suggest that our approach can enable a robot to predict and react to the forces a garment will exert onto a person during robot-assisted dressing. With a prediction horizon of 0.2s, our model predictive controller is able to fully dress a person's arm in clothing and mitigate the chance of a garment getting caught on a person's body.

\section{Conclusion}

In this work, we presented a learning-based MPC approach that allows a robot to predict the physical implications of its actions and reduce applied force to a person during robot-assisted dressing. We trained a recurrent model on data collected in a self-supervised setting from a physics-based dressing simulation. Unlike prior robot control approaches that use vision-based techniques, our model is able to predict the forces applied to a person's body using only haptic and kinematic measurements from a robot's end effector.

Our model is trained via purely supervised learning, which allows us to define a cost function for MPC post training. This cost function enables a robot to prioritize actions that minimize the predicted force applied to a person's body during physical assistance. Note that this cost function could be changed for different tasks or to allow for personalization, without needing to retrain the model. For a person with a weak or injured wrist, a new function might be defined that primarily focuses on reducing forces applied to the person's hand or wrist. When coupled with state estimation~\cite{yamazaki2013method,jimenez2017visual}, it may be possible to define dynamic cost functions that change depending on the current state of a task.

We evaluated our method with a PR2 that pulled the sleeve of a hospital gown onto the arms of 10 human participants. Our approach enables a robot to predict and react to the forces a garment will exert onto a person during robot-assisted dressing. Our approach also runs in real time on a PR2, using only the robot's on-board CPUs, yet computation time may be a limiting factor for tasks that require faster action replanning rates or longer prediction horizons. 
From our experiments, we observed emergent behaviors during dressing as we increased the prediction horizon for MPC. With a horizon of 0.2s, our predictive controller was able to fully dress a person's arm in clothing and mitigate the chance of the garment getting caught on the person's body. 



\section*{Acknowledgment}

\small
\textit{This work was supported by NSF award IIS-1514258 and AWS Cloud Credits for Research. Dr. Kemp is a cofounder, a board member, an equity holder, and the CTO of Hello Robot, Inc., which is developing products related to this research. This research could affect his personal financial status. The terms of this arrangement have been reviewed and approved by Georgia Tech in accordance with its conflict of interest policies.}

\bibliographystyle{IEEEtran}

\end{document}